\documentclass{article}

\usepackage[preprint]{arxiv_style}

\usepackage[utf8]{inputenc} 
\usepackage[T1]{fontenc}    
\usepackage{xurl}           
\usepackage{hyperref}       
\usepackage{booktabs}       
\usepackage{amsfonts}       
\usepackage{nicefrac}       
\usepackage{microtype}      
\usepackage{xcolor}         

\usepackage{amsmath}
\usepackage{enumitem}
\usepackage{colortbl}
\usepackage{graphicx} 
\usepackage{wrapfig}
\usepackage{placeins}
\usepackage{float}
\usepackage{pgfplots}
\pgfplotsset{compat=1.18}
\usepgfplotslibrary{groupplots}

\definecolor{highlightblue}{RGB}{232, 242, 255}
\definecolor{textblue}{RGB}{0, 51, 153}
\definecolor{titlegray}{gray}{0.65}

\newcommand{\PaperAuthorNames}{Shunya Nagashima$^{1}$, Takumi Bannai$^{2,3}$, Shuitsu Koyama$^{1}$, Tomoya Mitsui$^{1}$, Shuntaro Suzuki$^{1}$}
\newcommand{\PaperAffiliations}{$^{1}$Neurogica Inc. \quad $^{2}$LTS, Inc. \quad $^{3}$ME-Lab Japan, Inc.}

\makeatletter
\renewcommand{\@toptitlebar}{}
\renewcommand{\@bottomtitlebar}{\vskip 0.2in}
\renewcommand{\@noticestring}{}
\renewcommand{\@notice}{}
\renewcommand{\@maketitle}{%
  \vbox{%
    \hsize\textwidth
    \linewidth\hsize
    \vskip 0.02in
    {\color{titlegray}\hrule height 0.5pt}
    \vskip 1.1em
    {\raggedright\fontsize{24}{28}\selectfont\bfseries \@title\par}
    \vskip 1.0em
    {\raggedright\normalsize\bfseries \PaperAuthorNames\par}
    \vskip 0.55em
    {\raggedright\normalsize \PaperAffiliations\par}
    \vskip 1.5em
  }%
}
\renewenvironment{abstract}%
{%
  \vskip 0.3em%
  {\noindent\large\bfseries Abstract\par}%
  \vspace{0.5ex}%
  \noindent
}%
{%
  \par%
  \vskip 1.2ex%
}
\makeatother
 
\title{Neural Stochastic Processes \\ for Satellite Precipitation Refinement}

\author{\PaperAuthorNames}

\begin{document}

\maketitle

\begin{abstract}
Accurate precipitation estimation is critical for flood forecasting, water resource management, and disaster preparedness.
Satellite products provide global hourly coverage but contain systematic biases; ground-based gauges are accurate at point locations but too sparse for direct gridded correction.
Existing methods fuse these sources by interpolating gauge observations onto the satellite grid, but treat each time step independently and therefore discard temporal structure in precipitation fields.
We propose Neural Stochastic Process (NSP), a model that pairs a Neural Process encoder conditioning on arbitrary sets of gauge observations with a latent Neural SDE on a 2D spatial representation.
NSP is trained under a single variational objective with simulation-free cost.
We also introduce QPEBench, a benchmark of 43{,}756 hourly samples over the Contiguous United States (2021--2025) with four aligned data sources and six evaluation metrics.
On QPEBench, NSP outperforms 13 baselines across all six metrics and surpasses JAXA's operational gauge-calibrated product.
An additional experiment on Kyushu, Japan confirms generalization to a different region with independent data sources.
\end{abstract}

\section{Introduction}
Accurate precipitation monitoring underpins flood forecasting, water resource management, and disaster preparedness~\cite{alfieri2013glofas, loucks2017water, wmo2023atlas}; weather, climate, and water-related hazards imposed cumulative losses exceeding US\$4.3 trillion over 1970--2021~\cite{wmo2023atlas}.
Satellite products such as GSMaP~\cite{kubota2020gsmap} provide global hourly estimates but exhibit systematic biases~\cite{kidd2011satellite}, while ground-based radar coverage remains absent across most of Africa, central Asia, and oceanic regions~\cite{skofronick2017gpm}.
Correcting satellite biases with sparse gauge observations is therefore essential, especially where no ground-based alternative exists.

\begin{figure}[t]
    \centering
    \includegraphics[width=\linewidth]{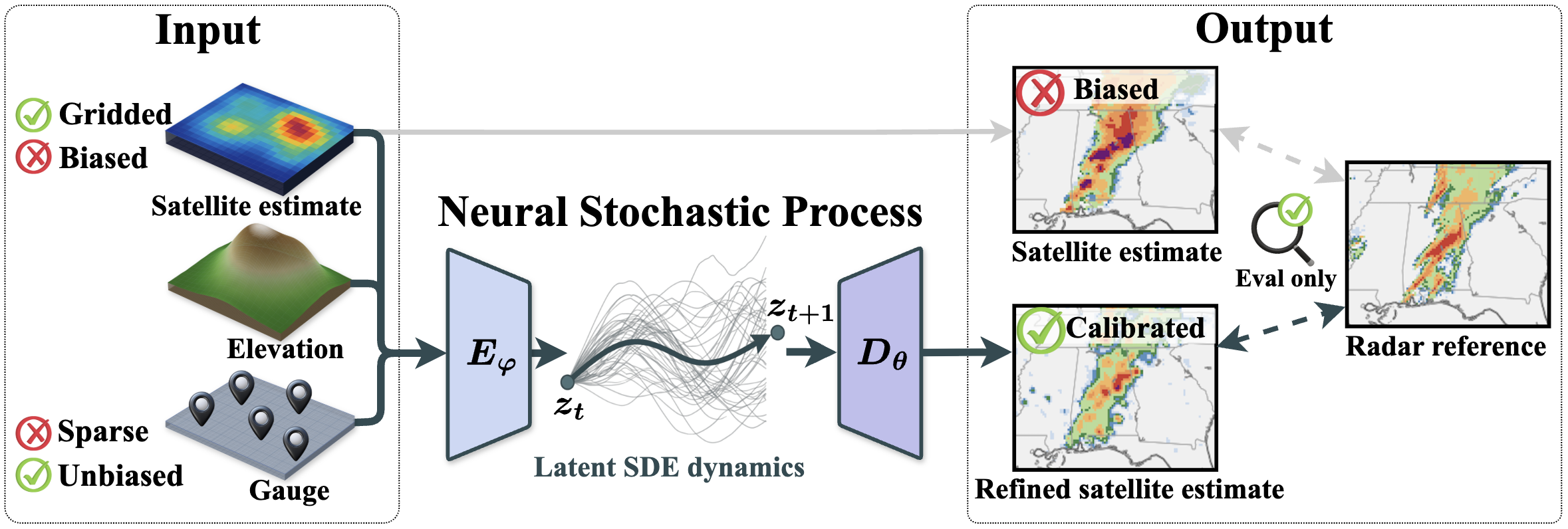}
    \vspace{-5mm}
    \caption{Motivation for the proposed Neural Stochastic Process (NSP).
    Satellite precipitation estimates provide gridded coverage but carry
    systematic biases, while gauges are sparse but locally accurate.
    NSP encodes satellite estimates, elevation, and gauges with the
    encoder $E_{\varphi}$, evolves a latent state through stochastic
    dynamics, and maps it to a calibrated precipitation estimate with
    the decoder $D_{\theta}$; ``Eval only'' indicates that the radar reference is
    used only for comparison, not as a model input.}
    \label{fig:eye-catch}
    \vspace{-3mm}
\end{figure}

In this paper, we address the refinement of satellite precipitation estimates by fusing spatially continuous satellite observations with highly accurate but spatially sparse point measurements from ground-based gauges to produce spatially coherent precipitation fields (Fig.~\ref{fig:eye-catch}).
The underlying problem—estimating a continuous spatio-temporal field from sparse, irregularly placed observations under non-stationary dynamics—has been studied through Gaussian processes~\cite{rasmussen2006gaussian}, neural processes~\cite{garnelo2018neural}, and neural differential equations~\cite{chen2018neural, li2020scalable}, yet spatial and temporal modeling remain largely disjoint.

This task is challenging because precipitation is zero-inflated and heavy-tailed~\cite{trenberth2017intermittency} and gauge observations are sparse and time-varying; furthermore, convective cells can evolve within an hour, making independent per-timestep estimation temporally incoherent.
Even JAXA's gauge-calibrated product achieved hourly correlation of only 0.31--0.47 against surface gauge observations in prior regional evaluations~\cite{lv2024gsmap}.

Existing approaches address only part of this challenge: geostatistical methods~\cite{cressie1993statistics, brunsdon1996gwr} are intractable at high resolution, pixel-wise corrections~\cite{shepard1968idw, wood2004qm, gneiting2005emos} ignore spatial structure, machine learning methods~\cite{chen2016xgboost, badrinath2023cnn} process each time step independently, and no standardized benchmark with unified evaluation exists for this task.

To address these limitations, we propose Neural Stochastic Process (NSP), a framework that unifies Neural Process-based spatial interpolation from sparse observations~\cite{garnelo2018neural} with Neural SDE-based continuous-time dynamics~\cite{li2020scalable} under a single variational objective.
NSP models joint spatio-temporal structure through a latent stochastic process on a 2D spatial field, producing temporally consistent predictions via a residual correction to the satellite estimate.

We also construct QPEBench (Quantitative Precipitation Estimation Benchmark), the first ML-ready benchmark for spatio-temporal precipitation refinement, covering five years (2021--2025) of hourly data over the Contiguous United States with radar-based reference fields and six metrics (Section~\ref{sec:qpebench}).

Our contributions are as follows:
\begin{itemize}[leftmargin=1.5em,itemsep=1pt,topsep=3pt]
\item A variational objective unifying Neural Process inference with Neural SDE dynamics; its temporal term is a closed-form transition KL from Girsanov's theorem~\cite{li2020scalable}, requiring no SDE solver calls.
\item A spatially-structured 2D latent field that replaces the global representation of standard NPs, enabling convolutional SDE dynamics with position-dependent uncertainty.
\item QPEBench, the first ML-ready benchmark for spatio-temporal precipitation refinement with six evaluation metrics covering accuracy and spatial structure.
\item A comprehensive empirical evaluation on QPEBench showing that NSP outperforms all 13 baselines and JAXA's operational gauge-calibrated product (GSMaP~GC) across all six metrics.
\end{itemize}

\vspace{-2mm}
\section{Related Work}
\vspace{-1mm}

\vspace{-0.5mm}
\paragraph{Satellite precipitation refinement.}
Satellite products such as GSMaP~\cite{kubota2020gsmap} and GPM IMERG~\cite{huffman2020imerg} provide global hourly estimates but exhibit systematic biases arising from indirect retrieval~\cite{kidd2011satellite}; comprehensive surveys are provided in~\cite{sun2018review_precip, maggioni2016review_satellite_precip}.
Classical methods range from geostatistical interpolation such as kriging~\cite{cressie1993statistics} to local bias-correction schemes such as IDW~\cite{shepard1968idw}, quantile mapping~\cite{wood2004qm}, and EMOS~\cite{gneiting2005emos}; they either rely on explicit spatial covariance modeling that is difficult to scale or adjust locations individually without producing spatially coherent precipitation fields.
CNN-based approaches have been applied to NWP post-processing~\cite{badrinath2023cnn} and satellite bias correction over complex terrain~\cite{chen2021deep}, though regression-based training tends to oversmooth heavy precipitation.
GANs produce stochastic precipitation fields that better capture the full intensity distribution~\cite{leinonen2020spagan}, and diffusion models conditioned on NWP or satellite variables demonstrate balanced skill across intensity quantiles~\cite{liu2025precipitation_diffusion, srivastava2024stvd}.
However, these methods typically assume dense conditioning variables or image-to-image settings rather than sparse, irregular gauge context, and none jointly models sparse-gauge interpolation with temporal dynamics.
We tested a conditional DDPM under our sparse-gauge protocol but found it unstable across folds; see Appendix~\ref{app:ddpm}.

\vspace{-0.5mm}
\paragraph{Precipitation and weather benchmarks.}
WeatherBench~2~\cite{rasp2024weatherbench2} and ChaosBench~\cite{nathaniel2024chaosbench} target forecast evaluation, while RainBench~\cite{dewitt2021rainbench}, RainNet~\cite{chen2022rainnet}, RainShift~\cite{harder2025rainshift}, and SatRain~\cite{pfreundschuh2025satrain} focus on precipitation forecasting, downscaling, or retrieval evaluation; see Appendix~\ref{app:related} for details.
To our knowledge, none of these benchmarks addresses fusing satellite estimates with sparse, irregularly distributed gauge observations.
QPEBench fills this gap with five years of aligned satellite, elevation, and gauge inputs alongside six metrics covering both accuracy and spatial structure.

\vspace{-0.5mm}
\paragraph{Neural Processes.}
Neural Processes (NPs)~\cite{garnelo2018conditional, garnelo2018neural} learn to predict dense outputs from sparse context observations in a single forward pass, providing a scalable alternative to Gaussian Processes; recent developments are surveyed in~\cite{bruinsma2024thesis}.
Subsequent work has introduced convolutional architectures that exploit spatial stationarity~\cite{gordon2020convolutional, foong2020convnp}, gridded transformer variants that scale to large spatio-temporal datasets~\cite{ashman2025gridtnp}, and flow matching for improved conditional generation~\cite{abuhamad2025flownp}; see Appendix~\ref{app:related} for a fuller catalogue.
The NP variants most relevant to this task still treat each time step independently.

\vspace{-0.5mm}
\paragraph{Neural SDEs.}
Neural ODEs~\cite{chen2018neural} parameterize continuous-time dynamics with neural networks; Neural SDEs~\cite{tzen2019neural, li2020scalable} generalize this by adding a learnable diffusion term, capturing both deterministic trends and stochastic fluctuations~\cite{kidger2022thesis}.
Training latent SDEs requires a path-level KL divergence that typically demands expensive simulation through SDE solvers~\cite{kidger2021efficient}.
Girsanov's theorem~\cite{li2020scalable} reduces this to a drift matching loss with no SDE solver calls, and recent work extends simulation-free training through score- and flow-matching formulations~\cite{bartosh2025sde}; see Appendix~\ref{app:related} for applications.
NSP applies convolutional drift and diffusion on a 2D spatial latent field, extending prior latent SDE formulations from vector-valued to spatially structured dynamics.

\vspace{-2mm}
\section{QPEBench}
\label{sec:qpebench}
\vspace{-1mm}

Existing ML-ready precipitation benchmarks address medium-range forecasting over gridded reanalysis fields~\cite{rasp2024weatherbench2, nathaniel2024chaosbench, dewitt2021rainbench} or paired low/high-resolution downscaling~\cite{chen2022rainnet, harder2025rainshift}, rather than fusing satellite estimates with sparse gauge observations.
We propose QPEBench, built from aligned multi-source inputs and an independent radar reference over the Contiguous United States (CONUS).

\vspace{-0.5mm}
\paragraph{Data sources.}
QPEBench combines four data sources on a common $260 \times 590$ grid at $0.1^\circ$ resolution (${\sim}11$\,km).
\emph{Satellite precipitation}: JAXA's GSMaP MVK product~\cite{kubota2020gsmap} provides gauge-uncorrected hourly estimates.
We chose GSMaP over NASA IMERG~\cite{huffman2020imerg} because the MVK (satellite-only) and GC (gauge-calibrated) variants share the same retrieval algorithm, enabling a controlled comparison of gauge fusion methods under the same satellite baseline.
\emph{Elevation}: ETOPO 2022~\cite{noaa2022etopo} supplies elevation, resampled to the $0.1^\circ$ grid.
\emph{Gauge observations}: hourly reports from 11{,}879 unique stations obtained through the Synoptic Data API~\cite{synopticdata}, totaling over 423 million records.
\emph{Radar reference}: NOAA's Multi-Radar Multi-Sensor (MRMS) Radar-Only QPE~\cite{smith2016mrms}, resampled to $0.1^\circ$, provides spatially dense, gauge-independent precipitation estimates and is \emph{strictly excluded from training}.
Models receive satellite, elevation, and gauge inputs; radar serves only as evaluation reference.

\vspace{-0.5mm}
\paragraph{Dataset statistics.}
The benchmark spans five years from January 2021 to December 2025, yielding 43{,}824 hourly time steps across diverse meteorological conditions including major hurricanes; see Appendix~\ref{app:dataset_stats}.
The spatial extent covers CONUS ($24^\circ$--$50^\circ$N, $125^\circ$W--$66^\circ$W), much larger than the regional domains used in prior QPE studies~\cite{yang2022correcting, baig2025bias}.
Hourly precipitation is zero-inflated and heavy-tailed: most pixels record no rain, yet the non-zero values span several orders of magnitude~\cite{trenberth2003changing}.
After quality filtering (see Appendix~\ref{app:filtering}), 43{,}756 samples remain, averaging 7{,}269 gauges per hour.

\vspace{-0.5mm}
\paragraph{Evaluation protocol.}
We employ three-fold expanding-window time-series cross-validation over the five-year period, advancing the test year from 2023 to 2025 while growing the training set chronologically to prevent temporal leakage; see Appendix~\ref{app:cv}.
Following standard QPE and spatial verification practice~\cite{ebert2008fuzzy, roberts2008fss}, QPEBench adopts six metrics covering both numerical accuracy and spatial structure:
$\text{RMSE}_r$ and $\text{MAE}_r$ evaluate the full gridded field against the independent radar reference;
$\text{RMSE}_g$ and $\text{MAE}_g$ quantify fidelity to gauge observations at discrete locations;
$r_{r,\text{coll}}$ measures the Pearson correlation between predictions and radar at locations where both gauge and radar exceed $0.1$\,mm/h, mitigating correlation inflation by the dominant dry class~\cite{tian2009component};
and $\text{FSS}_R$, the Fractions Skill Score averaged over four intensity thresholds (1.0, 2.5, 5.0, and 10.0\,mm/h) with a $20 \times 20$ pixel neighborhood, assesses whether the predicted field reproduces physically plausible precipitation patterns at the neighborhood scale.
Formal definitions are provided in Appendix~\ref{app:metrics}.

\vspace{-2mm}
\section{Neural Stochastic Process}
\vspace{-1mm}

Ground-based radar provides the most accurate precipitation fields but covers only a small fraction of the globe~\cite{skofronick2017gpm}; where radar is unavailable, satellite estimates must be refined using sparse, irregularly placed gauge observations whose count and locations change hourly.
Neural Processes~\cite{garnelo2018conditional, garnelo2018neural} are well suited to this setting: the encoder conditions on the subset of gauges available at each hour, treating them as context points that provide local bias supervision at observed locations, while the decoder generalizes to unobserved target locations, enabling a single model to accommodate the irregular, time-varying observation network without architectural modification.
We extend this framework with Neural SDEs~\cite{li2020scalable} under a unified variational objective, forming the \textbf{Neural Stochastic Process (NSP)} for spatio-temporal precipitation refinement.

\begin{figure*}[t]
    \centering
    \includegraphics[width=\linewidth]{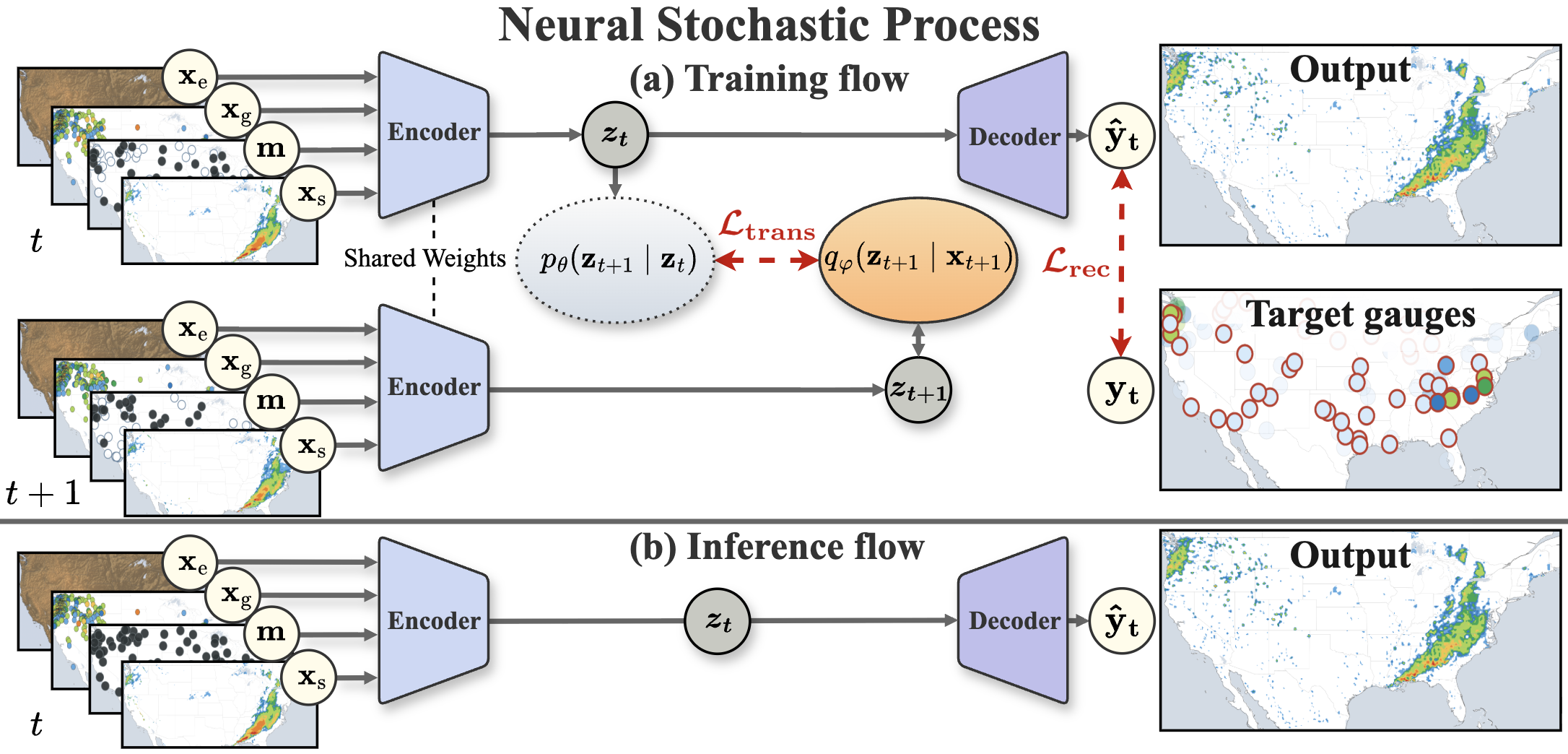}
    \vspace{-3mm}
    \caption{Architecture of Neural Stochastic Process (NSP).
    \textbf{(a)}~During training, gauge observations are split into context and target sets: the black markers in the input maps denote context gauges encoded through the mask $\mathbf{m}_t$, while the right panel shows target gauges $\mathbf{y}_t$ used by $\mathcal{L}_{\mathrm{rec}}$.
    The shared encoder maps consecutive inputs to $\mathbf{z}_t$ and $\mathbf{z}_{t+1}$, and $\mathcal{L}_{\mathrm{trans}}$ aligns the Neural SDE prior $p_\theta(\mathbf{z}_{t+1} \mid \mathbf{z}_t)$ with the encoder posterior $q_\varphi(\mathbf{z}_{t+1} \mid \mathbf{x}_{t+1})$.
    \textbf{(b)}~At inference, all available gauges are treated as context, and a single encoder--decoder pass predicts $\hat{\mathbf{y}}_t$ without SDE integration.}
    \label{fig:model}
\end{figure*}

NSP consists of three components (Fig.~\ref{fig:model}): an encoder mapping inputs to a spatially-structured latent distribution, a Neural SDE regularizing temporal dynamics, and a decoder predicting residual corrections.
At each time step $t$, the input $\mathbf{x}_t \in \mathbb{R}^{4 \times H \times W}$ consists of a satellite precipitation estimate $\mathbf{x}_{\mathrm{s},t} \in \mathbb{R}^{H \times W}$, elevation $\mathbf{x}_{\mathrm{e}} \in \mathbb{R}^{H \times W}$, sparse gauge observations $\mathbf{x}_{\mathrm{g},t} \in \mathbb{R}^{H \times W}$, and a binary mask $\mathbf{m}_t \in \{0, 1\}^{H \times W}$ indicating observed locations.

\vspace{-1mm}
\subsection{Context-to-target formulation}
\vspace{-1mm}

Following the Neural Process framework, gauge observations are partitioned during training into a context set $\mathcal{C}_t$ for the encoder and a target set $\mathcal{T}_t$ for the reconstruction loss.
We denote the target gauge observations by $\mathbf{y}_t = \{y_i : i \in \mathcal{T}_t\}$; at inference, all available gauges serve as context and $\hat{\mathbf{y}}_t$ is the full predicted precipitation field (partitioning details in Section~\ref{sec:setup}).
The encoder maps $\mathbf{x}_t$ to an approximate posterior over a spatially-structured latent state:
\begin{equation}
  q_\varphi(\mathbf{z}_t \mid \mathbf{x}_t)
  = \mathcal{N}\!\bigl(
    \boldsymbol{\mu}_\varphi(\mathbf{x}_t),\;
    \mathrm{diag}\bigl(\boldsymbol{\sigma}^2_\varphi(\mathbf{x}_t)\bigr)
  \bigr),
  \label{eq:encoder}
\end{equation}
where $\mathbf{z}_t \in \mathbb{R}^{D \times H' \times W'}$ preserves the spatial layout of $\mathbf{x}_t$ (Section~\ref{sec:spatial_latent}).
A sample $\mathbf{z}_t$ is drawn via the reparameterization trick~\cite{kingma2014auto}.
The decoder predicts a residual correction $\boldsymbol{\delta}_t$ in log-precipitation space and a predictive variance $\hat{\boldsymbol{\sigma}}^2_{y,t}$, conditioned on $\mathbf{z}_t$, $\mathbf{x}_{\mathrm{s},t}$, and $\mathbf{x}_{\mathrm{e}}$.
The refined estimate is
\begin{equation}
  \hat{\mathbf{y}}_t
  = \exp\!\bigl(
    \log(1 + \mathbf{x}_{\mathrm{s},t}) + \boldsymbol{\delta}_t
  \bigr) - 1,
  \label{eq:residual}
\end{equation}
and the reconstruction loss is evaluated at target locations against the held-out gauge observations.

\vspace{-1mm}
\subsection{Temporal dynamics via Neural SDE}
\vspace{-1mm}

Applying the Neural Process independently at each time step ignores temporal dependencies, leading to physically implausible discontinuities.
We model the latent temporal dynamics as an SDE:
\begin{equation}
  d\mathbf{z}_t = f_\theta(\mathbf{z}_t)\,dt
  + \boldsymbol{\sigma}_\theta(\mathbf{z}_t) \odot d\mathbf{W}_t,
  \label{eq:prior_sde}
\end{equation}
where $f_\theta$ and $\boldsymbol{\sigma}_\theta$ are neural networks parameterizing the drift and element-wise diffusion (Section~\ref{sec:spatial_latent}), $\odot$ denotes the Hadamard product, $\mathbf{W}_t$ is a Wiener process, and $\boldsymbol{\sigma}_\theta$ is constrained to be positive.

\vspace{-1mm}
\subsection{Unified variational objective}
\label{sec:objective}
\vspace{-1mm}

A heuristic combination of independent Neural Process losses with an auxiliary SDE penalty lacks a clear probabilistic interpretation.
The objective is to retain the deployed single-hour encoder while encouraging temporally consistent latent trajectories.
Accordingly, we use a factorized approximate posterior for per-hour inference and introduce temporal coupling through the SDE transition term.
We derive a single variational bound that decomposes into a spatial reconstruction term (the standard NP ELBO) and a temporal drift-matching term (the Neural SDE transition loss), thereby unifying the two frameworks under one generative model.

\vspace{-0.5mm}
\paragraph{Generative model.}
The joint distribution over observations $\mathbf{y}_{0:T}$ and latent states $\mathbf{z}_{0:T}$ factorizes as
\begin{equation}
  p_\theta(\mathbf{y}_{0:T}, \mathbf{z}_{0:T})
  = p(\mathbf{z}_0)
  \prod_{t=0}^{T-1} p_\theta(\mathbf{z}_{t+1} \mid \mathbf{z}_t)
  \prod_{t=0}^{T} p_\theta(\mathbf{y}_t \mid \mathbf{z}_t),
  \label{eq:joint}
\end{equation}
where $p(\mathbf{z}_0) = \mathcal{N}(\mathbf{0}, \mathbf{I})$, $p_\theta(\mathbf{z}_{t+1} \mid \mathbf{z}_t)$ is the SDE transition, and $p_\theta(\mathbf{y}_t \mid \mathbf{z}_t)$ denotes the decoder likelihood; the decoder additionally conditions on the satellite field $\mathbf{x}_{\mathrm{s},t}$ and elevation $\mathbf{x}_{\mathrm{e}}$, which we suppress from the notation as they are always observed.

\vspace{-0.5mm}
\paragraph{Proposition 1 (ELBO decomposition).}
We introduce a mean-field approximate posterior $q_\varphi(\mathbf{z}_{0:T} \mid \mathbf{x}_{0:T}) = \prod_t q_\varphi(\mathbf{z}_t \mid \mathbf{x}_t)$, where each factor is given by~(\ref{eq:encoder}).
Unlike latent SDE formulations that use temporally coupled posterior inference~\cite{li2020scalable, kidger2021efficient}, this approximation keeps encoder inference parallel across time and consistent with test time, where each $\mathbf{z}_t$ must be inferred from a single hour's observation alone.
A joint posterior over the full spatio-temporal latent field would require sequential inference over all time steps and is substantially more expensive in our high-dimensional 2D setting.
Temporal coupling is then introduced through $\mathcal{L}_{\mathrm{trans}}$, which aligns the independently encoded $q_\varphi(\mathbf{z}_{t+1} \mid \mathbf{x}_{t+1})$ with the SDE transition from $\mathbf{z}_t$; Section~\ref{sec:results} and Appendix~\ref{app:temporal_ablation} show that removing this term degrades gridded accuracy.
Factorizing the joint distribution~(\ref{eq:joint}) and applying Jensen's inequality yields the following bound; see Appendix~\ref{app:elbo}:
\begin{align}
  -\log p_\theta(\mathbf{y}_{0:T})
  \;\leq\;&
  \underbrace{
    -\sum_{t=0}^{T} \mathbb{E}_{q_\varphi(\mathbf{z}_t \mid \mathbf{x}_t)}\!\bigl[
      \log p_\theta(\mathbf{y}_t \mid \mathbf{z}_t)
    \bigr]
  }_{\mathcal{L}_{\mathrm{rec}}}
  \;+\;
  \underbrace{
    D_{\mathrm{KL}}\!\bigl(
      q_\varphi(\mathbf{z}_0 \mid \mathbf{x}_0)
      \,\|\, p(\mathbf{z}_0)
    \bigr)
  }_{\mathcal{L}_{\mathrm{prior}}}
  \notag \\[4pt]
  &+\;
  \underbrace{
    \sum_{t=0}^{T-1}
    \mathbb{E}_{q_\varphi(\mathbf{z}_t \mid \mathbf{x}_t)}\!\Bigl[
      D_{\mathrm{KL}}\!\bigl(
        q_\varphi(\mathbf{z}_{t+1} \mid \mathbf{x}_{t+1})
        \,\|\, p_\theta(\mathbf{z}_{t+1} \mid \mathbf{z}_t)
      \bigr)
    \Bigr]
  }_{\mathcal{L}_{\mathrm{trans}}},
  \label{eq:elbo}
\end{align}
where $p_\theta(\mathbf{y}_t \mid \mathbf{z}_t)$ is a heteroscedastic Gaussian whose mean and variance are produced by the decoder.
$\mathcal{L}_{\mathrm{rec}}$ is the standard Neural Process reconstruction loss summed over all time steps, and $\mathcal{L}_{\mathrm{prior}}$ regularizes the initial latent state $\mathbf{z}_0$ toward the standard Gaussian prior; together they recover the single-step NP ELBO at $t{=}0$.
$\mathcal{L}_{\mathrm{trans}}$ couples consecutive time steps by comparing two distributions over $\mathbf{z}_{t+1}$: the encoder's posterior $q_\varphi(\mathbf{z}_{t+1} \mid \mathbf{x}_{t+1})$, obtained by independently encoding the next observation $\mathbf{x}_{t+1}$, and the SDE's forward prediction $p_\theta(\mathbf{z}_{t+1} \mid \mathbf{z}_t)$ from the current latent state.
Minimizing this divergence trains $f_\theta$ to approximate the encoder's latent transition from $\mathbf{z}_t$ alone.

\vspace{-0.5mm}
\paragraph{Proposition 2 (solver-free transition KL and Girsanov reduction).}
Under an Euler--Maruyama discretization of~(\ref{eq:prior_sde}) with step size $\Delta t$, the SDE transition becomes Gaussian:
\begin{equation}
  p_\theta(\mathbf{z}_{t+1} \mid \mathbf{z}_t)
  = \mathcal{N}\!\bigl(
    \mathbf{z}_t + f_\theta(\mathbf{z}_t)\,\Delta t,\;
    \mathrm{diag}(\boldsymbol{\sigma}_\theta(\mathbf{z}_t)^2)\,\Delta t
  \bigr).
  \label{eq:euler}
\end{equation}
Since both $q_\varphi(\mathbf{z}_{t+1} \mid \mathbf{x}_{t+1})$ and $p_\theta(\mathbf{z}_{t+1} \mid \mathbf{z}_t)$ are diagonal Gaussians, $\mathcal{L}_{\mathrm{trans}}$ admits a closed-form, simulation-free evaluation with no SDE solver calls, avoiding the expensive numerical integration required by adjoint-based latent SDE training~\cite{li2020scalable}.
When the encoder variance $\boldsymbol{\sigma}^2_\varphi$ is constrained to equal $\boldsymbol{\sigma}_\theta^2 \Delta t$, $\mathcal{L}_{\mathrm{trans}}$ reduces to the Girsanov drift matching loss~\cite{li2020scalable}; Appendix~\ref{app:temporal_ablation} compares this special case and a naive temporal penalty against the full transition KL.

\vspace{-0.5mm}
\paragraph{Training objective.}
The decoder likelihood $p_\theta(\mathbf{y}_t \mid \mathbf{z}_t)$ is a heteroscedastic Gaussian evaluated at target locations $\mathcal{T}_t$.
The final training loss augments the bound~(\ref{eq:elbo}) with two auxiliary terms that have natural probabilistic interpretations: $\mathcal{L}_{\mathrm{ctx}}$ applies the same NLL at context locations $\mathcal{C}_t$, and $\mathcal{L}_{\delta} = \sum_{t=0}^{T}\|\boldsymbol{\delta}_t\|^2$ can be interpreted as an $\ell_2$ regularizer, equivalently a zero-mean Gaussian prior on the residual correction.
We further add $\mathcal{L}_{\mathrm{prior}} = \sum_{t=0}^{T} D_{\mathrm{KL}}\!\bigl(q_\varphi(\mathbf{z}_t \mid \mathbf{x}_t)\,\|\,\mathcal{N}(\mathbf{0}, \mathbf{I})\bigr)$, extending the $t{=}0$ KL regularization to all time steps as an auxiliary stabilizer.
The resulting objective is
\begin{equation}
  \mathcal{L}_{\mathrm{NSP}}
  = \mathcal{L}_{\mathrm{rec}}
  + \beta_{\mathrm{ctx}}\,\mathcal{L}_{\mathrm{ctx}}
  + \beta_{\mathrm{kl}}\,\mathcal{L}_{\mathrm{prior}}
  + \beta_{\mathrm{sde}}\,\mathcal{L}_{\mathrm{trans}}
  + \beta_{\delta}\,\mathcal{L}_{\delta},
  \label{eq:total_loss}
\end{equation}
where $\beta_{\mathrm{kl}}$, $\beta_{\mathrm{sde}}$, $\beta_{\mathrm{ctx}}$, and $\beta_{\delta}$ are weighting coefficients.
We apply stop-gradient to $\mathbf{z}_t$ when computing $\mathcal{L}_{\mathrm{trans}}$, decoupling the encoder's spatial inference from the SDE's temporal dynamics.

\vspace{-0.5mm}
\paragraph{Inference.}
At test time, the encoder maps $\mathbf{x}_t$ to $q_\varphi(\mathbf{z}_t \mid \mathbf{x}_t)$, a sample $\mathbf{z}_t$ is drawn, and the decoder produces $\hat{\mathbf{y}}_t$ via~(\ref{eq:residual}); samples yield pixel-level uncertainty estimates.
The SDE is not rolled forward at inference; instead, $\mathcal{L}_{\mathrm{trans}}$ acts as a training-time temporal regularizer, and each hourly prediction requires only a single encoder--decoder pass without sequential SDE integration.

\vspace{-1mm}
\subsection{Spatially-structured 2D latent field}
\label{sec:spatial_latent}
\vspace{-1mm}

Standard Neural Processes collapse all context observations into a global $\mathbf{z} \in \mathbb{R}^D$~\cite{garnelo2018neural}, losing spatial correlations and forcing all positions to share a single source of stochasticity~\cite{kohl2018probabilistic}.
Convolutional variants~\cite{gordon2020convolutional} retain spatial structure but employ deterministic inference without temporal dynamics.

NSP instead uses a 2D latent feature map $\mathbf{z}_t \in \mathbb{R}^{D \times H' \times W'}$, where $H' = H / r$, $W' = W / r$, and $r$ is the encoder's downsampling ratio.
The drift $f_\theta$ and diffusion $\boldsymbol{\sigma}_\theta$ use $3 {\times} 3$ convolutions on $\mathbf{z}_t$, so that temporal dynamics at each position are informed by its spatial neighborhood.
The decoder upsamples $\mathbf{z}_t$ to full resolution, fusing the satellite field and elevation to predict $\boldsymbol{\delta}_t$ and $\hat{\boldsymbol{\sigma}}^2_{y,t}$.

\vspace{-2mm}
\section{Experimental Setup}
\label{sec:setup}
\vspace{-1mm}

\paragraph{Dataset.}
We used QPEBench (Section~\ref{sec:qpebench}) with its four-channel input and radar evaluation reference, yielding 43{,}756 usable hourly samples after quality filtering.
Three-fold expanding-window cross-validation (Section~\ref{sec:qpebench}) was used to select the model with lowest validation loss.
We employed the six metrics defined in Section~\ref{sec:qpebench} and Appendix~\ref{app:metrics}, jointly assessing gridded accuracy, gauge-level fidelity, and spatial coherence.

\vspace{-0.5mm}
\paragraph{Baselines.}
We compared NSP against 13 baselines: twelve methods across three groups---geostatistical interpolation (IDW, kriging, cokriging, GWR)~\cite{shepard1968idw,cressie1993statistics,brunsdon1996gwr}, statistical post-processing (quantile mapping, EMOS, linear regression)~\cite{wood2004qm,gneiting2005emos}, and machine learning (XGBoost, U-Net, CNP, ConvCNP, ViT)~\cite{chen2016xgboost,ronneberger2015unet,garnelo2018conditional,gordon2020convolutional,dosovitskiy2021vit}---along with raw GSMaP as an input-quality reference~\cite{kubota2020gsmap}.
All machine learning baselines received four-channel input and were trained under an identical context/target protocol for fair comparison; training objectives and the ConvCNP reimplementation are detailed in Appendix~\ref{app:baseline_loss}.
We also include GSMaP~GC, JAXA's operational gauge-calibrated product that adjusts satellite-only estimates using ${\sim}$11{,}500 CPC gauges (${\sim}$26\,km mean spacing over CONUS) with a three-day latency~\cite{mega2019gauge, kubota2020gsmap}.
Although the total Synoptic station count available to NSP is comparable (${\sim}$11{,}900), the mean hourly availability is ${\sim}$7{,}300 stations (${\sim}$33\,km spacing), making NSP's effective input sparser.

\vspace{-0.5mm}
\paragraph{Implementation details.}
NSP has 4.19M parameters ($D{=}64$, hidden width 128).
At each training step, 50\% of gauges were randomly assigned to the context set and the rest to targets, with balanced rainy/non-rainy sampling; at test time all gauges served as context.
Training used AdamW with OneCycleLR (peak $3{\times}10^{-3}$) for 4 epochs on eight NVIDIA H200 GPUs (effective batch size 32), completing in approximately 20 minutes across all three folds.
Inference required 2.64~ms per sample.
Full hyperparameters are provided in Appendix~\ref{app:hyperparams}.

\vspace{-0.5mm}
\paragraph{Additional regional experiment.}
To evaluate geographic generalizability, we conducted an additional experiment over the Kyushu region of Japan with independent data sources; model and training are identical to CONUS (setup in Appendix~\ref{app:kyushu}; results in Section~\ref{sec:results}).

\begin{table*}[!t]
    \centering
    \renewcommand{\arraystretch}{1.25}
    \caption{Quantitative comparison of precipitation refinement methods. Values are mean $\pm$ standard deviation over three-fold time-series cross-validation. The best score in each column is in \textbf{bold}.}
    \vspace{1mm}
    \label{tab:performance_comparison_refined}
    \resizebox{\textwidth}{!}{%
    \begin{tabular}{@{} l c c c c c c @{}}
    \toprule
    \textbf{Method} & $\text{RMSE}_r \downarrow$ & $\text{MAE}_r \downarrow$ & $\text{RMSE}_g \downarrow$ & $\text{MAE}_g \downarrow$ & $r_{r,\text{coll}} \uparrow$ & $\text{FSS}_R \uparrow$ \\
    \midrule
    Quantile mapping~\cite{wood2004qm}   & 4.073{\scriptsize $\pm$0.184} & 2.012{\scriptsize $\pm$0.063} & 0.885{\scriptsize $\pm$0.027} & 0.173{\scriptsize $\pm$0.006} & 0.288{\scriptsize $\pm$0.014} & 0.479{\scriptsize $\pm$0.010} \\
    EMOS~\cite{gneiting2005emos}               & 3.885{\scriptsize $\pm$0.127} & 1.844{\scriptsize $\pm$0.078} & 1.115{\scriptsize $\pm$0.096} & 0.179{\scriptsize $\pm$0.019} & 0.305{\scriptsize $\pm$0.013} & 0.478{\scriptsize $\pm$0.029} \\
    XGBoost~\cite{chen2016xgboost}            & 3.741{\scriptsize $\pm$0.129} & 1.827{\scriptsize $\pm$0.077} & 1.062{\scriptsize $\pm$0.088} & 0.179{\scriptsize $\pm$0.020} & 0.284{\scriptsize $\pm$0.016} & 0.487{\scriptsize $\pm$0.008} \\
    GWR~\cite{brunsdon1996gwr}                & 3.739{\scriptsize $\pm$0.230} & 1.627{\scriptsize $\pm$0.081} & 0.662{\scriptsize $\pm$0.021} & 0.155{\scriptsize $\pm$0.004} & 0.376{\scriptsize $\pm$0.016} & 0.229{\scriptsize $\pm$0.017} \\
    Cokriging~\cite{cressie1993statistics}          & 3.706{\scriptsize $\pm$0.121} & 1.789{\scriptsize $\pm$0.069} & 1.065{\scriptsize $\pm$0.081} & 0.220{\scriptsize $\pm$0.015} & 0.279{\scriptsize $\pm$0.016} & 0.461{\scriptsize $\pm$0.009} \\
    GSMaP~\cite{kubota2020gsmap}              & 3.638{\scriptsize $\pm$0.126} & 1.826{\scriptsize $\pm$0.076} & 1.017{\scriptsize $\pm$0.086} & 0.179{\scriptsize $\pm$0.020} & 0.288{\scriptsize $\pm$0.014} & 0.483{\scriptsize $\pm$0.014} \\
    Kriging~\cite{cressie1993statistics}            & 3.204{\scriptsize $\pm$0.076} & 1.645{\scriptsize $\pm$0.027} & 0.774{\scriptsize $\pm$0.022} & 0.177{\scriptsize $\pm$0.005} & 0.006{\scriptsize $\pm$0.003} & 0.007{\scriptsize $\pm$0.002} \\
    ConvCNP~\cite{gordon2020convolutional}      & 3.135{\scriptsize $\pm$0.098} & 1.596{\scriptsize $\pm$0.046} & 0.684{\scriptsize $\pm$0.018} & 0.100{\scriptsize $\pm$0.003} & 0.475{\scriptsize $\pm$0.005} & 0.053{\scriptsize $\pm$0.011} \\
    U-Net~\cite{ronneberger2015unet}            & 3.123{\scriptsize $\pm$0.057} & 1.563{\scriptsize $\pm$0.038} & 0.666{\scriptsize $\pm$0.028} & 0.100{\scriptsize $\pm$0.004} & 0.461{\scriptsize $\pm$0.021} & 0.064{\scriptsize $\pm$0.052} \\
    CNP~\cite{garnelo2018conditional}                & 3.115{\scriptsize $\pm$0.077} & 1.575{\scriptsize $\pm$0.056} & 0.699{\scriptsize $\pm$0.033} & 0.116{\scriptsize $\pm$0.006} & 0.340{\scriptsize $\pm$0.018} & 0.014{\scriptsize $\pm$0.011} \\
    ViT~\cite{dosovitskiy2021vit}                & 3.059{\scriptsize $\pm$0.108} & 1.530{\scriptsize $\pm$0.079} & 0.687{\scriptsize $\pm$0.022} & 0.116{\scriptsize $\pm$0.016} & 0.411{\scriptsize $\pm$0.003} & 0.118{\scriptsize $\pm$0.019} \\
    IDW~\cite{shepard1968idw}                & 2.987{\scriptsize $\pm$0.070} & 1.474{\scriptsize $\pm$0.020} & 0.645{\scriptsize $\pm$0.018} & 0.145{\scriptsize $\pm$0.004} & 0.352{\scriptsize $\pm$0.010} & 0.157{\scriptsize $\pm$0.002} \\
    Linear regression  & 2.967{\scriptsize $\pm$0.064} & 1.457{\scriptsize $\pm$0.020} & 0.709{\scriptsize $\pm$0.019} & 0.170{\scriptsize $\pm$0.005} & 0.299{\scriptsize $\pm$0.017} & 0.189{\scriptsize $\pm$0.012} \\
    \midrule
    \rowcolor{gray!15}
    GSMaP GC~\cite{mega2019gauge}           & 2.942{\scriptsize $\pm$0.085} & 1.473{\scriptsize $\pm$0.039} & 0.737{\scriptsize $\pm$0.039} & 0.147{\scriptsize $\pm$0.007} & 0.375{\scriptsize $\pm$0.022} & 0.490{\scriptsize $\pm$0.023} \\
    \rowcolor{highlightblue}
    \textbf{NSP (Ours)} & \textbf{2.818}{\scriptsize $\pm$0.062} & \textbf{1.444}{\scriptsize $\pm$0.026} & \textbf{0.393}{\scriptsize $\pm$0.047} & \textbf{0.076}{\scriptsize $\pm$0.013} & \textbf{0.478}{\scriptsize $\pm$0.021} & \textbf{0.527}{\scriptsize $\pm$0.022} \\
    \bottomrule
    \end{tabular}
    }
\end{table*}

\vspace{-2mm}
\section{Results}
\label{sec:results}
\vspace{-1mm}

\subsection{Quantitative Comparison}
\begin{figure*}[t]
    \centering
    \includegraphics[width=\linewidth]{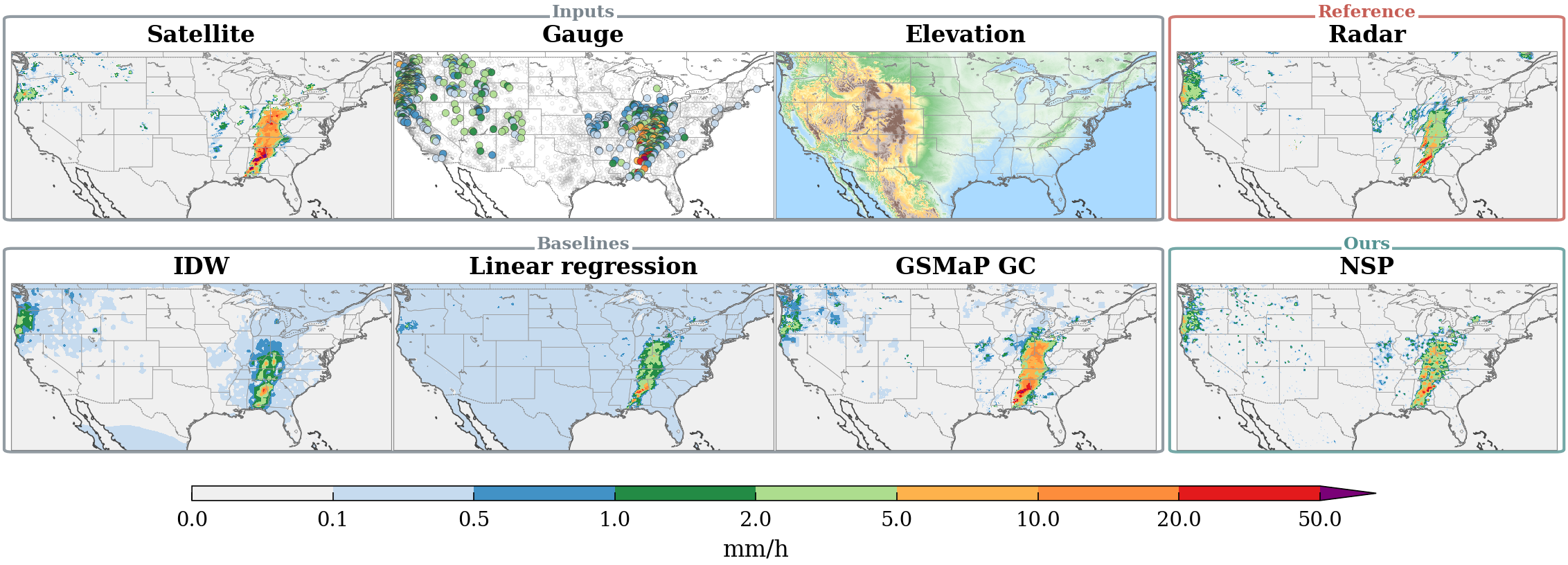}
    \vspace{-3mm}
    \caption{Qualitative result at 03:00~UTC on March~16, 2025.
    NSP better matches the radar reference despite the satellite overestimation.}
    \label{fig:qualitative}
    \vspace{-3mm}
\end{figure*}
\begin{figure}[t]
    \centering
    \includegraphics[width=\linewidth]{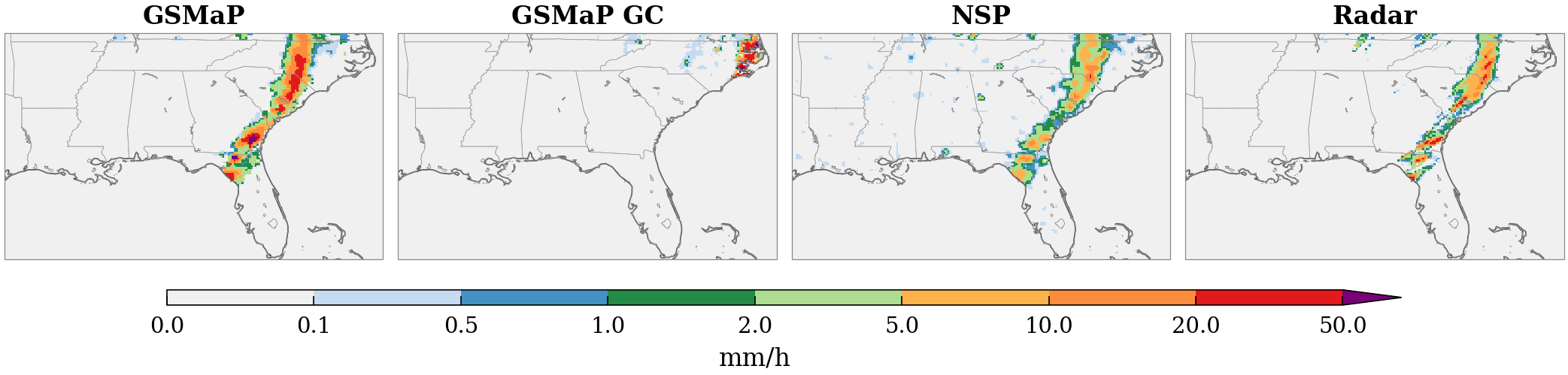}
    \vspace{-5mm}
    \caption{Regional zoom over the southeastern United States at 16:00~UTC on March~16, 2025.}
    \label{fig:regional}
    \vspace{-3mm}
\end{figure}

Table~\ref{tab:performance_comparison_refined} shows the comparison on CONUS.
NSP achieved the best performance across all six metrics, outperforming all 13 baselines as well as JAXA's operational gauge-calibrated product (GSMaP~GC) despite sparser effective gauge input (${\sim}$7{,}300 hourly stations vs.\ GC's ${\sim}$11{,}500; Section~\ref{sec:setup}).
Against the second-best method on each metric, NSP reduced $\text{RMSE}_r$ by 4.2\% (2.818 vs.\ GSMaP~GC's 2.942), $\text{RMSE}_g$ by 39.1\% (0.393 vs.\ IDW's 0.645), and $\text{FSS}_R$ by 7.6\% relative improvement (0.527 vs.\ GSMaP~GC's 0.490).
U-Net, CNP, ViT, and ConvCNP achieved competitive gauge-level accuracy ($\text{RMSE}_g$: 0.666, 0.699, 0.687, and 0.684) but much lower $\text{FSS}_R$ (0.064, 0.014, 0.118, and 0.053), showing that these methods overfit to gauge locations at the expense of spatial structure, whereas NSP preserves spatial structure without sacrificing point-level accuracy.

\FloatBarrier
\vspace{-0.5mm}
\paragraph{Kyushu regional results.}
\begin{wraptable}{r}{0.46\textwidth}
\vspace{-18pt}
\centering
\footnotesize
\caption{Kyushu (Japan) regional results.}
\label{tab:kyushu}
\setlength{\tabcolsep}{4pt}
\renewcommand{\arraystretch}{1.05}
\begin{tabular}{@{} l ccc @{}}
\toprule
\textbf{Method} & $\text{RMSE}_r\!\downarrow$ & $\text{RMSE}_g\!\downarrow$ & $r_{r,\text{coll}}\!\uparrow$ \\
\midrule
GWR~\cite{brunsdon1996gwr}              & 1.788 & 0.655 & 0.389 \\
CNP~\cite{garnelo2018conditional}       & 1.779 & 1.953 & 0.437 \\
EMOS~\cite{gneiting2005emos}            & 1.738 & 1.082 & 0.169 \\
Cokriging~\cite{cressie1993statistics}   & 1.715 & 1.077 & 0.139 \\
Linear regression & 1.688 & 0.768 & 0.356 \\
Kriging~\cite{cressie1993statistics}     & 1.640 & 0.973 & 0.114 \\
GSMaP~\cite{kubota2020gsmap}            & 1.508 & 0.965 & 0.158 \\
XGBoost~\cite{chen2016xgboost}          & 1.501 & 0.985 & 0.190 \\
ConvCNP~\cite{gordon2020convolutional}  & 1.363 & 0.645 & 0.525 \\
U-Net~\cite{ronneberger2015unet}        & 1.324 & 0.679 & 0.532 \\
IDW~\cite{shepard1968idw}               & 1.272 & 0.680 & 0.397 \\
\midrule
\rowcolor{gray!15}
GSMaP GC~\cite{mega2019gauge}           & 1.288 & 0.762 & 0.265 \\
\rowcolor{highlightblue}
\textbf{NSP}     & \textbf{1.119} & \textbf{0.485} & \textbf{0.553} \\
\bottomrule
\end{tabular}
\vspace{-10pt}
\end{wraptable}

Table~\ref{tab:kyushu} reports the additional experiment on the Kyushu region of Japan (Section~\ref{sec:setup}), which features a different precipitation regime (East Asian monsoon) and higher gauge density than CONUS.
NSP achieved the best performance across all three reported metrics, reducing $\text{RMSE}_r$ by 13.1\% relative to GSMaP~GC (1.119 vs.\ 1.288) and $\text{RMSE}_g$ by 36.4\% (0.485 vs.\ 0.762).
The correlation $r_{r,\text{coll}}$ improved from 0.265 to 0.553, a 2.1$\times$ gain, supporting cross-domain generalization.
Among baselines, IDW achieved the second-lowest $\text{RMSE}_r$ (1.272) and ViT the second-highest $r_{r,\text{coll}}$ (0.543), reflecting the same accuracy--coherence trade-off as on CONUS.
$\text{FSS}_R$ is omitted (small grid); full results are in Appendix~\ref{app:kyushu}.

\vspace{-2mm}
\subsection{Qualitative Results}
\vspace{-1mm}

To assess NSP's performance under extreme conditions, we visualize predictions from the March 14--16, 2025 severe weather outbreak, the largest March tornado event on record with over 180 confirmed tornadoes and more than \$10~billion in damage~\citep{ncei2025billions}.
Fig.~\ref{fig:qualitative} presents a full CONUS sample: the satellite estimate overestimates precipitation across the central United States, while IDW produces scattered artifacts and GSMaP~GC retains the satellite bias.
NSP corrects the overestimation using gauge observations and elevation, producing a field that agrees well with the radar reference.
Fig.~\ref{fig:regional} zooms into the southeastern United States: GSMaP overestimates the spatial extent of precipitation, GSMaP~GC markedly attenuates the signal, while NSP recovers a narrow, well-localized precipitation band consistent with the radar reference.

\vspace{-1mm}
\subsection{Ablation Study}
\vspace{-1mm}
\begin{table*}[t]
\centering
\renewcommand{\arraystretch}{1.15}
\caption{Ablation study on CONUS. Mean $\pm$ std over three folds; best in \textbf{bold}.}
\label{tab:ablation}
\resizebox{\textwidth}{!}{%
\begin{tabular}{@{} cl cccccc @{}}
\toprule
& \textbf{Model} & $\text{RMSE}_r \downarrow$ & $\text{MAE}_r \downarrow$ & $\text{RMSE}_g \downarrow$ & $\text{MAE}_g \downarrow$ & $r_{r,\text{coll}} \uparrow$ & $\text{FSS}_R \uparrow$ \\
\midrule
\multicolumn{8}{@{}l}{\emph{Architecture ablations}} \\
(1-i)   & w/o Neural SDE             & 2.888{\scriptsize $\pm$0.039} & 1.487{\scriptsize $\pm$0.053} & 0.338{\scriptsize $\pm$0.066} & 0.062{\scriptsize $\pm$0.018} & 0.479{\scriptsize $\pm$0.020} & 0.509{\scriptsize $\pm$0.022} \\
(1-ii)  & Deterministic              & 2.885{\scriptsize $\pm$0.080} & 1.492{\scriptsize $\pm$0.044} & 0.355{\scriptsize $\pm$0.147} & 0.060{\scriptsize $\pm$0.028} & \textbf{0.487}{\scriptsize $\pm$0.033} & 0.495{\scriptsize $\pm$0.012} \\
(1-iii) & w/o Satellite residual     & 3.141{\scriptsize $\pm$0.089} & 1.613{\scriptsize $\pm$0.038} & \textbf{0.252}{\scriptsize $\pm$0.024} & \textbf{0.041}{\scriptsize $\pm$0.005} & 0.475{\scriptsize $\pm$0.018} & 0.153{\scriptsize $\pm$0.047} \\
(1-iv)  & Homoscedastic              & 2.932{\scriptsize $\pm$0.062} & 1.492{\scriptsize $\pm$0.020} & 0.260{\scriptsize $\pm$0.016} & 0.052{\scriptsize $\pm$0.003} & 0.471{\scriptsize $\pm$0.023} & 0.445{\scriptsize $\pm$0.012} \\
(1-v)   & Log-space MSE              & 3.554{\scriptsize $\pm$0.104} & 1.726{\scriptsize $\pm$0.048} & 0.405{\scriptsize $\pm$0.031} & 0.043{\scriptsize $\pm$0.006} & 0.475{\scriptsize $\pm$0.019} & 0.525{\scriptsize $\pm$0.022} \\
\midrule
\multicolumn{8}{@{}l}{\emph{Loss ablations}} \\
(2-i)   & w/o $\mathcal{L}_{\mathrm{trans}}$  & 2.932{\scriptsize $\pm$0.069} & 1.524{\scriptsize $\pm$0.054} & 0.306{\scriptsize $\pm$0.008} & 0.053{\scriptsize $\pm$0.003} & 0.477{\scriptsize $\pm$0.020} & 0.517{\scriptsize $\pm$0.032} \\
(2-ii)  & w/o $\mathcal{L}_{\mathrm{prior}}$  & 2.969{\scriptsize $\pm$0.123} & 1.551{\scriptsize $\pm$0.050} & 0.278{\scriptsize $\pm$0.034} & 0.045{\scriptsize $\pm$0.007} & 0.475{\scriptsize $\pm$0.020} & 0.514{\scriptsize $\pm$0.019} \\
(2-iii) & w/o $\mathcal{L}_{\mathrm{ctx}}$    & 2.994{\scriptsize $\pm$0.025} & 1.610{\scriptsize $\pm$0.036} & 0.819{\scriptsize $\pm$0.040} & 0.154{\scriptsize $\pm$0.012} & 0.303{\scriptsize $\pm$0.017} & 0.479{\scriptsize $\pm$0.029} \\
\midrule
\rowcolor{highlightblue}
& \textbf{NSP (Full)}        & \textbf{2.818}{\scriptsize $\pm$0.062} & \textbf{1.444}{\scriptsize $\pm$0.026} & 0.393{\scriptsize $\pm$0.047} & 0.076{\scriptsize $\pm$0.013} & 0.478{\scriptsize $\pm$0.021} & \textbf{0.527}{\scriptsize $\pm$0.022} \\
\bottomrule
\end{tabular}
}
\vspace{-5pt}
\end{table*}

Table~\ref{tab:ablation} isolates each component's contribution.
The full model achieved the best $\text{RMSE}_r$, $\text{MAE}_r$, and $\text{FSS}_R$; several variants improved gauge-level metrics at the cost of spatial coherence, showing that naively minimizing gauge error overfits the sparse observation network.
Removing the satellite residual connection (1-iii) collapsed $\text{FSS}_R$ from 0.527 to 0.153 despite yielding the lowest $\text{RMSE}_g$: without the satellite prior, the model degenerates into gauge interpolation and loses gridded structure.
Replacing the heteroscedastic NLL with MSE (1-v) degraded $\text{RMSE}_r$ by 26.1\%, showing that learned variance is essential for handling zero-inflated precipitation.

Among the loss components, removing $\mathcal{L}_{\mathrm{ctx}}$ (2-iii) severely degraded gauge accuracy ($\text{RMSE}_g$: 0.393$\to$0.819), showing that context reconstruction is necessary for stable training.
Removing $\mathcal{L}_{\mathrm{trans}}$ (2-i) or $\mathcal{L}_{\mathrm{prior}}$ (2-ii) each increased $\text{RMSE}_r$ by 4--6\%, showing that both the SDE transition and KL regularization prevent overfitting to per-time-step gauge signals. This supports the factorized posterior design in Section~\ref{sec:objective}: although the encoder infers each hour independently, $\mathcal{L}_{\mathrm{trans}}$ reintroduces the temporal coupling needed for improved gridded accuracy. Appendix~\ref{app:temporal_ablation} further shows that the full transition KL provides the most favorable trade-off relative to both a naive temporal penalty and the matched-variance special case.
Appendix~\ref{app:event_temporal} tracks two 12-hour fold~3 events hour by hour, where NSP outperformed the no-$\mathcal{L}_{\mathrm{trans}}$ ablation in 22/24 hourly steps in $\mathrm{RMSE}_r$ and 21/24 in $\mathrm{FSS}_R$; Appendix~\ref{app:sde_rollout} shows that pure SDE rollout remains competitive only over short horizons, supporting our choice to use the SDE mainly as a training-time temporal regularizer.

\vspace{-2mm}
\subsection{Error Analysis}
\vspace{-1mm}
\begin{table*}[t]
\centering
\footnotesize
\caption{Error analysis. \textbf{(a)}~Pixel-level RMSE by radar precipitation intensity (fold~3, 8{,}759 samples). \textbf{(b)}~Median spatial displacement (km) between predicted and radar exceedance masks, measured by the shift maximizing 2-D cross-correlation; lower is better.}
\label{tab:error_analysis}
\vspace{1pt}
\setlength{\tabcolsep}{3pt}
\renewcommand{\arraystretch}{1.1}
\begin{tabular}[t]{@{} l ccc rr @{}}
\multicolumn{6}{c}{\textbf{(a) RMSE by intensity}}\\[4pt]
\toprule
\textbf{Intensity} & \smash{\raisebox{-4.5pt}{\textbf{GSMaP}}} & \smash{\raisebox{-4.5pt}{\textbf{GC}}} & \smash{\raisebox{-4.5pt}{\textbf{NSP}}} & \multicolumn{2}{c}{\textbf{Rel.\ improv.}} \\
\cmidrule(l){5-6}
\textbf{(mm/h)} & & & & \textbf{vs.\ GSMaP} & \textbf{vs.\ GC} \\
\midrule
0          & 0.592 & 0.377 & \textbf{0.308} & $+$48.0\% & $+$18.3\% \\
0--1       & 0.693 & 0.436 & \textbf{0.356} & $+$48.6\% & $+$18.3\% \\
1--5       & 4.011 & 2.528 & \textbf{2.055} & $+$48.8\% & $+$18.7\% \\
5--10      & 7.725 & 5.597 & \textbf{5.190} & $+$32.8\% & $+$7.3\%  \\
10--25     & 13.684 & 12.250 & \textbf{12.152} & $+$11.2\% & $+$0.8\%  \\
25+        & \textbf{30.037} & 29.847 & 30.086 & $-$0.2\% & $-$0.8\%  \\
\bottomrule
\end{tabular}
\hfill
\begin{tabular}[t]{@{} l ccc rr @{}}
\multicolumn{6}{c}{\textbf{(b) Spatial displacement (km)}}\\[4pt]
\toprule
\textbf{Thresh.} & \smash{\raisebox{-4.5pt}{\textbf{GSMaP}}} & \smash{\raisebox{-4.5pt}{\textbf{GC}}} & \smash{\raisebox{-4.5pt}{\textbf{NSP}}} & \multicolumn{2}{c}{\textbf{Rel.\ improv.}} \\
\cmidrule(l){5-6}
\textbf{(mm/h)} & & & & \textbf{vs.\ GSMaP} & \textbf{vs.\ GC} \\
\midrule
0.1  & 11  & 11  & 11          &    0\% &    0\% \\
1.0  & 16  & 16  & \textbf{11} & $+$29\% & $+$29\% \\
5.0  & 40  & 45  & \textbf{31} & $+$21\% & $+$31\% \\
10.0 & \textbf{66}  & 113 & 70 & $-$7\%  & $+$38\% \\
\bottomrule
\end{tabular}
\end{table*}

Table~\ref{tab:error_analysis}(a) presents pixel-level RMSE stratified by radar precipitation intensity.
NSP outperformed raw GSMaP by 33--49\% and GSMaP~GC by 7--19\% for intensities up to 10\,mm/h; at extreme precipitation ($>$25\,mm/h), all methods converged, indicating that heavy-rain estimation remains a shared bottleneck.
Table~\ref{tab:error_analysis}(b) reports exceedance-mask displacement. NSP reduces GSMaP~GC displacement by 29--38\% at 1--10\,mm/h, indicating better spatial alignment of rainy areas. At 10\,mm/h, raw GSMaP still shows slightly smaller displacement (66\,km vs.\ 70\,km), suggesting that the remaining gap is more likely due to intensity estimation than large spatial mislocalization.
Per-sample failure decomposition and extended error analysis are in Appendix~\ref{app:error}; see Appendix~\ref{app:gauge_sensitivity} for gauge density sensitivity, which shows that NSP maintains spatial accuracy even at 10\% context density.

\FloatBarrier
\vspace{-2mm}
\section{Conclusion}
\vspace{-1mm}

We proposed NSP, a Neural Process coupled with a latent Neural SDE for satellite precipitation refinement.
The model operates on a 2D spatial latent field and is trained with a simulation-free drift matching loss from Girsanov's theorem with no SDE solver calls.
We also introduced QPEBench, a five-year, hourly, CONUS-scale benchmark with an independent radar reference and six evaluation metrics.
On QPEBench, NSP outperformed 13 baselines on all six metrics and surpassed JAXA's operational gauge-calibrated product; a separate experiment on Kyushu, Japan showed consistent gains under different data sources and gauge density.

\vspace{-1mm}
\paragraph{Limitations.}
NSP underestimates heavy rain due to the zero-inflated training distribution.
QPEBench covers only CONUS; radar-sparse regions will require gauge-only evaluation.
The SDE loss encourages but does not enforce temporal consistency at inference.

\vspace{-1mm}
\paragraph{Future work.}
Future work includes intensity-aware sampling or tail-weighted losses for heavy-rain reconstruction, and broader validation in radar-sparse regions such as sub-Saharan Africa.

\bibliography{main}

@String(NeurIPS = {Advances in Neural Information Processing Systems (NeurIPS)})

@String(ICML  = {Proceedings of the International Conference on Machine Learning (ICML)})

@String(ICLR    = {International Conference on Learning Representations (ICLR)})

@String(AAAI    = {Proceedings of the AAAI Conference on Artificial Intelligence (AAAI)})

@article{alfieri2013glofas,
  title={{GloFAS}--global ensemble streamflow forecasting and flood early warning},
  author={Alfieri, Lorenzo and Burek, Peter and Dutra, Emanuel and Krzeminski, Blazej and Muraro, David and Thielen, Jutta and Pappenberger, Florian},
  journal={Hydrology and Earth System Sciences},
  volume={17},
  number={3},
  pages={1161--1175},
  year={2013},
  publisher={Copernicus Publications G{\"o}ttingen, Germany}
}

@book{loucks2017water,
  title={Water resource systems planning and management: An introduction to methods, models, and applications},
  author={Loucks, Daniel P and Van Beek, Eelco},
  year={2017},
  publisher={Springer}
}

@misc{wmo2023atlas,
  author={{World Meteorological Organization}},
  title={Atlas of Mortality and Economic Losses from Weather, Climate and Water-related Hazards (1970--2021)},
  year={2023},
  month=may,
  howpublished={\url{https://wmo.int/publication-series/atlas-of-mortality-and-economic-losses-from-weather-climate-and-water-related-hazards-1970-2021}},
  note={Published 22 May 2023; accessed 2026-03-18}
}

@article{skofronick2017gpm,
  title={The Global Precipitation Measurement ({GPM}) mission for science and society},
  author={Skofronick-Jackson, Gail and Petersen, Walter A and Berg, Wesley and Kidd, Chris and Stocker, Erich F and Kirschbaum, Dalia B and Kakar, Ramesh and others},
  journal={Bulletin of the American Meteorological Society},
  volume={98},
  number={8},
  pages={1679--1695},
  year={2017}
}

@incollection{kubota2020gsmap,
  title={Global Satellite Mapping of Precipitation ({GSMaP}) products in the {GPM} era},
  author={Kubota, Takuji and Aonashi, Kazumasa and Ushio, Tomoo and Shige, Shoichi and Takayabu, Yukari N and Kachi, Misako and Arai, Yoriko and others},
  booktitle={Satellite precipitation measurement: Volume 1},
  pages={355--373},
  year={2020},
  publisher={Springer}
}

@article{kidd2011satellite,
  title={Status of satellite precipitation retrievals},
  author={Kidd, Chris and Levizzani, Vincenzo},
  journal={Hydrology and Earth System Sciences},
  volume={15},
  number={4},
  pages={1109--1116},
  year={2011},
  publisher={Copernicus Publications G{\"o}ttingen, Germany}
}

@article{trenberth2003changing,
  title={The changing character of precipitation},
  author={Trenberth, Kevin E and Dai, Aiguo and Rasmussen, Roy M and others},
  journal={Bulletin of the American Meteorological Society},
  volume={84},
  number={9},
  pages={1205--1218},
  year={2003},
  publisher={American Meteorological Society}
}

@book{cressie1993statistics,
  title={Statistics for spatial data},
  author={Cressie, Noel},
  year={1993},
  publisher={John Wiley \& Sons}
}

@inproceedings{chen2016xgboost,
  title={Xgboost: A scalable tree boosting system},
  author={Chen, Tianqi and Guestrin, Carlos},
  booktitle={Proceedings of the ACM SIGKDD International Conference on Knowledge Discovery and Data Mining},
  pages={785--794},
  year={2016}
}

@article{garnelo2018neural,
  title={Neural processes},
  author={Garnelo, Marta and Schwarz, Jonathan and Rosenbaum, Dan and Viola, Fabio and Rezende, Danilo J and Eslami, SM and Teh, Yee Whye},
  journal={arXiv preprint arXiv:1807.01622},
  year={2018}
}

@article{mega2019gauge,
  author={Mega, Tomoaki and Ushio, Tomoo and Takahiro, Matsuda and Kubota, Takuji and Kachi, Misako and Oki, Riko},
  journal={IEEE Transactions on Geoscience and Remote Sensing}, 
  title={Gauge-Adjusted Global Satellite Mapping of Precipitation}, 
  year={2019},
  volume={57},
  number={4},
  pages={1928--1935},
}

@article{sun2018review_precip,
  title={A review of global precipitation data sets: Data sources, estimation, and intercomparisons},
  author={Sun, Qiaohong and Miao, Chiyuan and Duan, Qingyun and Ashouri, Hamed and Sorooshian, Soroosh and Hsu, Kuo-Lin},
  journal={Reviews of Geophysics},
  volume={56},
  number={1},
  pages={79--107},
  year={2018},
  publisher={Wiley Online Library}
}

@article{maggioni2016review_satellite_precip,
  title={A review of merged high-resolution satellite precipitation product accuracy during the Tropical Rainfall Measuring Mission (TRMM) era},
  author={Maggioni, Viviana and Meyers, Patrick C and Robinson, Monique D},
  journal={Journal of Hydrometeorology},
  volume={17},
  number={4},
  pages={1101--1117},
  year={2016}
}

@inproceedings{garnelo2018conditional,
  title={Conditional neural processes},
  author={Garnelo, Marta and Rosenbaum, Dan and Maddison, Christopher and Ramalho, Tiago and Saxton, David and Shanahan, Murray and Teh, Yee Whye and Rezende, Danilo and others},
  booktitle=ICML,
  pages={1704--1713},
  year={2018},
  organization={PMLR}
}

@article{bruinsma2024thesis,
  title={Convolutional Conditional Neural Processes},
  author={Bruinsma, Wessel P},
  journal={arXiv preprint arXiv:2408.09583},
  year={2024}
}

@inproceedings{li2020scalable,
  title={Scalable gradients for stochastic differential equations},
  author={Li, Xuechen and Wong, Ting-Kam Leonard and Chen, Ricky TQ and Duvenaud, David},
  booktitle={International Conference on Artificial Intelligence and Statistics (AISTATS)},
  pages={3870--3882},
  year={2020},
  organization={PMLR}
}

@book{rasmussen2006gaussian,
  title={Gaussian Processes for Machine Learning},
  author={Rasmussen, Carl Edward and Williams, Christopher K. I.},
  publisher={MIT Press},
  year={2006}
}

@article{chen2018neural,
  title={Neural ordinary differential equations},
  author={Chen, Ricky TQ and Rubanova, Yulia and Bettencourt, Jesse and others},
  journal=NeurIPS,
  volume={31},
  year={2018}
}

@article{kidger2022thesis,
  title={On neural differential equations},
  author={Kidger, Patrick},
  journal={arXiv preprint arXiv:2202.02435},
  year={2022}
}

@incollection{huffman2020imerg,
  title={Integrated multi-satellite retrievals for the global precipitation measurement ({GPM}) mission ({IMERG})},
  author={Huffman, George J and Bolvin, David T and Braithwaite, Dan and Hsu, Kuo-Lin and Joyce, Robert J and others},
  booktitle={Satellite Precipitation Measurement: Volume 1},
  pages={343--353},
  year={2020},
  publisher={Springer}
}

@inproceedings{shepard1968idw,
  title={A two-dimensional interpolation function for irregularly-spaced data},
  author={Shepard, Donald},
  booktitle={Proceedings of the ACM National Conference},
  pages={517--524},
  year={1968}
}

@article{wood2004qm,
  title={Hydrologic implications of dynamical and statistical approaches to downscaling climate model outputs},
  author={Wood, Andrew W and Leung, Lai R and Sridhar, Venkataramana and Lettenmaier, Dennis P},
  journal={Climatic change},
  volume={62},
  number={1},
  pages={189--216},
  year={2004},
  publisher={Springer}
}

@article{gneiting2005emos,
  title={Calibrated probabilistic forecasting using ensemble model output statistics and minimum CRPS estimation},
  author={Gneiting, Tilmann and Raftery, Adrian E and Westveld III, Anton H and Goldman, Tom},
  journal={Monthly weather review},
  volume={133},
  number={5},
  pages={1098--1118},
  year={2005}
}

@article {badrinath2023cnn,
  title={Improving Precipitation Forecasts with Convolutional Neural Networks},
  author={Anirudhan Badrinath and Luca Delle Monache and Negin Hayatbini and Will Chapman and Forest Cannon and Marty Ralph},
  journal={Weather and Forecasting},
  year={2023},
  publisher={American Meteorological Society},
  address={Boston MA, USA},
  volume={38},
  number={2},
  doi={10.1175/WAF-D-22-0002.1},
  pages={291--306},
  url={https://journals.ametsoc.org/view/journals/wefo/38/2/WAF-D-22-0002.1.xml}
}

@article{chen2021deep,
  title={Deep learning for bias correction of satellite retrievals of orographic precipitation},
  author={Chen, Haonan and Sun, Luyao and Cifelli, Robert and Xie, Pingping},
  journal={IEEE Transactions on Geoscience and Remote Sensing},
  volume={60},
  number={},
  pages={1--11},
  year={2022},
  publisher={IEEE}
}

@article{leinonen2020spagan,
  title={Stochastic super-resolution for downscaling time-evolving atmospheric fields with a generative adversarial network},
  author={Leinonen, Jussi and Nerini, Daniele and Berne, Alexis},
  journal={IEEE Transactions on Geoscience and Remote Sensing},
  volume={59},
  number={9},
  pages={7211--7223},
  year={2021},
  publisher={IEEE}
}

@article{liu2025precipitation_diffusion,
  title={Precipitation estimation with NWP model and generative diffusion model},
  author={Liu, Haolin and Fung, Jimmy CH and Lau, Alexis KH and Li, Zhenning},
  journal={Geophysical Research Letters},
  volume={52},
  number={7},
  pages={e2024GL110625},
  year={2025},
  publisher={Wiley Online Library}
}

@inproceedings{chen2022rainnet,
  title={{RainNet}: A Large-Scale Imagery Dataset and Benchmark for Spatial Precipitation Downscaling},
  author={Chen, Xuanhong and Feng, Kairui and Liu, Naiyuan and Ni, Bingbing and Lu, Yifan and others},
  booktitle={Advances in Neural Information Processing Systems},
  volume={35},
  pages={15152--15164},
  year={2022}
}

@article{harder2025rainshift,
  title={{RainShift}: A Benchmark for Precipitation Downscaling Across Geographies},
  author={Harder, Paula and Schmidt, Luca and Pelletier, Francis and Ludwig, Nicole and Chantry, Matthew and Lessig, Christian and Hernandez-Garcia, Alex and Rolnick, David},
  journal={arXiv preprint arXiv:2507.04930},
  year={2025}
}

@inproceedings{dewitt2021rainbench,
  title={{RainBench}: Towards data-driven global precipitation forecasting from satellite imagery},
  author={de Witt, Christian Schroeder and Tong, Catherine and Zantedeschi, Valentina and others},
  booktitle=AAAI,
  volume={35},
  pages={14902--14910},
  year={2021}
}

@inproceedings{kim2019anp,
  title={Attentive Neural Processes},
  author={Hyunjik Kim and Andriy Mnih and Jonathan Schwarz and Marta Garnelo and Ali Eslami and Dan Rosenbaum and Oriol Vinyals and Yee Whye Teh},
  booktitle=ICLR,
  year={2019},
  url={https://openreview.net/forum?id=SkE6PjC9KX},
}

@inproceedings{gordon2020convolutional,
  title={Convolutional Conditional Neural Processes},
  author={Jonathan Gordon and Wessel P. Bruinsma and Andrew Y. K. Foong and James Requeima and Yann Dubois and Richard E. Turner},
  booktitle=ICLR,
  year={2020},
  url={https://openreview.net/forum?id=Skey4eBYPS}
}

@article{foong2020convnp,
  title={Meta-learning stationary stochastic process prediction with convolutional neural processes},
  author={Foong, Andrew and Bruinsma, Wessel and Gordon, Jonathan and Dubois, Yann and Requeima, James and others},
  journal=NeurIPS,
  volume={33},
  pages={8284--8295},
  year={2020}
}

@article{vaughan2022convnp_climate,
  title={Convolutional conditional neural processes for local climate downscaling},
  author={Vaughan, Anna and Tebbutt, Will and others},
  journal={Geoscientific Model Development},
  volume={15},
  number={1},
  pages={251--268},
  year={2022},
  publisher={Copernicus GmbH}
}

@article{allen2025end,
  title={End-to-end data-driven weather prediction},
  author={Allen, Anna and Markou, Stratis and Tebbutt, Will and Requeima, James and Bruinsma, Wessel P and Andersson, Tom R and Herzog, Michael and Lane, Nicholas D and Chantry, Matthew and Hosking, J Scott and others},
  journal={Nature},
  volume={641},
  number={8065},
  pages={1172--1179},
  year={2025},
  publisher={Nature Publishing Group UK London}
}

@inproceedings{bruinsma2023arcnp,
  title={Autoregressive Conditional Neural Processes},
  author={Wessel Bruinsma and Stratis Markou and James Requiema and Andrew Y. K. Foong and Tom Andersson and Anna Vaughan and Anthony Buonomo and Scott Hosking and others},
  booktitle=ICLR,
  year={2023},
  url={https://openreview.net/forum?id=OAsXFPBfTBh}
}

@article{tzen2019neural,
  title={Neural stochastic differential equations: Deep latent gaussian models in the diffusion limit},
  author={Tzen, Belinda and Raginsky, Maxim},
  journal={arXiv preprint arXiv:1905.09883},
  year={2019}
}

@article{kidger2021efficient,
  title={Efficient and accurate gradients for neural sdes},
  author={Kidger, Patrick and Foster, James and Li, Xuechen Chen and Lyons, Terry},
  journal=NeurIPS,
  volume={34},
  pages={18747--18761},
  year={2021}
}

@inproceedings{bartosh2025sde,
  title={{SDE} Matching: Scalable and Simulation-Free Training of Latent Stochastic Differential Equations},
  author={Bartosh, Grigory and Vetrov, Dmitry and others},
  booktitle=ICML,
  volume={267},
  pages={3054--3070},
  year={2025},
  organization={PMLR}
}

@inproceedings{oh2024stable,
  title={Stable Neural Stochastic Differential Equations in Analyzing Irregular Time Series Data},
  author={YongKyung Oh and Dongyoung Lim and Sungil Kim},
  booktitle=ICLR,
  year={2024},
  url={https://openreview.net/forum?id=4VIgNuQ1pY}
}

@inproceedings{ansari2023neural,
  title={Neural continuous-discrete state space models for irregularly-sampled time series},
  author={Ansari, Abdul Fatir and Heng, Alvin and Lim, Andre and Soh, Harold},
  booktitle=ICML,
  volume={202},
  pages={926--951},
  year={2023},
  organization={PMLR}
}

@inproceedings{zhang2024neural,
  title={Neural Jump-Diffusion Temporal Point Processes},
  author={Shuai Zhang and Chuan Zhou and Yang Aron Liu and Peng Zhang and Xixun Lin and Zhi-Ming Ma},
  booktitle=ICML,
  year={2024},
  volume={235},
  pages={60541--60557},
  organization={PMLR}
}

@inproceedings{dosovitskiy2021vit,
  title={An Image is Worth 16x16 Words: Transformers for Image Recognition at Scale},
  author={Alexey Dosovitskiy and Lucas Beyer and Alexander Kolesnikov and Dirk Weissenborn and Xiaohua Zhai and Thomas Unterthiner and others},
  booktitle=ICLR,
  year={2021},
  url={https://openreview.net/forum?id=YicbFdNTTy}
}

@misc{synopticdata,
  author       = {{Synoptic Data PBC}},
  title        = {Synoptic Weather API},
  year         = {2026},
  howpublished = {Data access platform},
  note         = {Accessed for station observations across the CONUS domain}
}

@article{smith2016mrms,
  title={Multi-Radar Multi-Sensor ({MRMS}) severe weather and aviation products: Initial operating capabilities},
  author={Smith, Travis M and Lakshmanan, Valliappa and Stumpf, Gregory J and Ortega, Kiel L and Hondl, Kurt and others},
  journal={Bulletin of the American Meteorological Society},
  volume={97},
  number={9},
  pages={1617--1630},
  year={2016}
}

@article{yang2022correcting,
  title={Correcting the bias of daily satellite precipitation estimates in tropical regions using deep neural network},
  author={Yang, Xiaoying and Yang, Shuai and Tan, Mou Leong and Pan, Hengyang and Zhang, Hongliang and Wang, Guoqing and He, Ruimin and Wang, Zimeng},
  journal={Journal of Hydrology},
  volume={608},
  pages={127656},
  year={2022},
  publisher={Elsevier}
}

@article{baig2025bias,
  title={From bias to accuracy: Transforming satellite precipitation data in arid regions with machine learning and topographical insights},
  author={Baig, Faisal and Ali, Luqman and Faiz, Muhammad Abrar and Chen, Haonan and Sherif, Mohsen},
  journal={Journal of Hydrology},
  volume={653},
  pages={132801},
  year={2025},
  publisher={Elsevier}
}

@article{ebert2008fuzzy,
  title={Fuzzy verification of high-resolution gridded forecasts: a review and proposed framework},
  author={Ebert, Elizabeth E},
  journal={Meteorological Applications: a Journal of Forecasting, Practical Applications, Training Techniques and Modelling},
  volume={15},
  number={1},
  pages={51--64},
  year={2008},
  publisher={Wiley Online Library}
}

@article{roberts2008fss,
  title={Scale-selective verification of rainfall accumulations from high-resolution forecasts of convective events},
  author={Roberts, Nigel M and others},
  journal={Monthly Weather Review},
  volume={136},
  number={1},
  pages={78--97},
  year={2008}
}

@article{kohl2018probabilistic,
  title={A probabilistic u-net for segmentation of ambiguous images},
  author={Kohl, Simon and Romera-Paredes, Bernardino and Meyer, Clemens and De Fauw, Jeffrey and Ledsam, Joseph R and Maier-Hein, Klaus and Eslami, SM and others},
  journal=NeurIPS,
  volume={31},
  year={2018}
}

@misc{noaa2022etopo,
  title={{ETOPO} 2022 15 arc-second global relief model},
  author={NOAA National Centers for Environmental Information},
  journal={NOAA National Centers for Environmental Information},
  year={2022}
}

@article{brunsdon1996gwr,
  title={Geographically weighted regression: a method for exploring spatial nonstationarity},
  author={Brunsdon, Chris and Fotheringham, A Stewart and others},
  journal={Geographical analysis},
  volume={28},
  number={4},
  pages={281--298},
  year={1996},
  publisher={Wiley Online Library}
}

@article{trenberth2017intermittency,
  title={Intermittency in Precipitation: Duration, Frequency, Intensity, and Amounts Using Hourly Data},
  author={Kevin E. Trenberth and others},
  journal={Journal of Hydrometeorology},
  year={2017},
  publisher={American Meteorological Society},
  volume={18},
  number={5},
  pages={1393-1412},
  url={https://journals.ametsoc.org/view/journals/hydr/18/5/jhm-d-16-0263_1.xml}
}

@article{lv2024gsmap,
  title={Evaluation of {GSMaP} Version 8 Precipitation Products on an Hourly Timescale over Mainland {China}},
  author={Lv, Xiaoyu and Guo, Hao and Tian, Yunfei and Meng, Xiangchen and Bao, Anming and De Maeyer, Philippe},
  journal={Remote Sensing},
  volume={16},
  number={1},
  pages={210},
  year={2024},
  doi={10.3390/rs16010210}
}

@article{srivastava2024stvd,
  title={Precipitation downscaling with spatiotemporal video diffusion},
  author={Srivastava, Prakhar and Yang, Ruihan and Kerrigan, Gavin and Dresdner, Gideon and McGibbon, Jeremy and Bretherton, Christopher and others},
  journal=NeurIPS,
  volume={37},
  pages={56374--56400},
  year={2024}
}

@article{abuhamad2025flownp,
title={Flow Matching Neural Processes},
author={Hussen Abu Hamad and Dan Rosenbaum},
journal=NeurIPS,
year={2025},
pages={1--24}
}

@inproceedings{
verma2024climode,
title={Clim{ODE}: Climate and Weather Forecasting with Physics-informed Neural {ODE}s},
author={Yogesh Verma and others},
booktitle=ICLR,
year={2024},
url={https://openreview.net/forum?id=xuY33XhEGR}
}

@inproceedings{ashman2025gridtnp,
  title={Gridded Transformer Neural Processes for Spatio-Temporal Data},
  author={Ashman, Matthew and Diaconu, Cristiana and Langezaal, Eric and Weller, Adrian and others},
  booktitle=ICML,
  pages={1722--1761},
  year={2025},
  volume={267},
  organization={PMLR},
}

@article{
rasp2024weatherbench2,
author = {Rasp, Stephan and Hoyer, Stephan and Merose, Alexander and Langmore, Ian and Battaglia, Peter and Russell, Tyler and others},
title = {{WeatherBench 2}: A Benchmark for the Next Generation of Data-Driven Global Weather Models},
journal = {Journal of Advances in Modeling Earth Systems},
volume = {16},
number = {6},
pages = {e2023MS004019},
doi = {https://doi.org/10.1029/2023MS004019},
url = {https://agupubs.onlinelibrary.wiley.com/doi/abs/10.1029/2023MS004019},
eprint = {https://agupubs.onlinelibrary.wiley.com/doi/pdf/10.1029/2023MS004019},
year = {2024}
}

@article{nathaniel2024chaosbench,
  title={{ChaosBench}: A multi-channel, physics-based benchmark for subseasonal-to-seasonal climate prediction},
  author={Nathaniel, Juan and Qu, Yongquan and Nguyen, Tung and Yu, Sungduk and Busecke, Julius and others},
  journal=NeurIPS,
  volume={37},
  pages={43715--43729},
  year={2024}
}

@article{pfreundschuh2025satrain,
  title={A Benchmark Dataset for Satellite-Based Estimation and Detection of Rain},
  author={Pfreundschuh, Simon and Arulraj, Malarvizhi and Behrangi, Ali and Bogerd, Linda and Calheiros, Alan James Peixoto and Casella, Daniele and Dolatabadi, Neda and Guilloteau, Clement and others},
  journal={Scientific Data},
  volume={13},
  number={1},
  pages={244},
  year={2026},
  doi={10.1038/s41597-026-06565-0}
}

@article{mohseni2025specconvcnp,
title={Spectral Convolutional Conditional Neural Process},
author={Peiman Mohseni and Nick Duffield},
journal=NeurIPS,
year={2025},
pages={1--31}
}

@inproceedings{park2025acssm,
title={Amortized Control of Continuous State Space Feynman-Kac Model for Irregular Time Series},
author={Byoungwoo Park and Hyungi Lee and Juho Lee},
booktitle=ICLR,
year={2025},
url={https://openreview.net/forum?id=8zJRon6k5v}
}

@article{kingma2014auto,
  title={Auto-encoding variational {B}ayes},
  author={Kingma, Diederik P and Welling, Max},
  journal={arXiv preprint arXiv:1312.6114},
  year={2014}
}

@article{tian2009component,
author = {Tian, Yudong and Peters-Lidard, Christa D. and Eylander, John B. and Joyce, Robert J. and Huffman, George J. and Adler, Robert F. and Hsu, Kuo-lin and others},
title = {Component analysis of errors in satellite-based precipitation estimates},
journal = {Journal of Geophysical Research: Atmospheres},
volume = {114},
number = {D24},
pages = {},
doi = {https://doi.org/10.1029/2009JD011949},
url = {https://agupubs.onlinelibrary.wiley.com/doi/abs/10.1029/2009JD011949},
eprint = {https://agupubs.onlinelibrary.wiley.com/doi/pdf/10.1029/2009JD011949},
year = {2009}
}

@article{pathak2022fourcastnet,
  title={FourCastNet: A Global Data-driven High-resolution Weather Model using Adaptive Fourier Neural Operators},
  author={Pathak, Jaideep and Subramanian, Shashank and Harrington, Peter and Raja, Sanjeev and Chattopadhyay, Ashesh and Mardani, Morteza and Kurth, Thorsten and others},
  journal={arXiv preprint arXiv:2202.11214},
  year={2022}
}

@article{bi2023pangu,
  title={Accurate medium-range global weather forecasting with {3D} neural networks},
  author={Bi, Kaifeng and Xie, Lingxi and Zhang, Hengheng and Chen, Xin and Gu, Xiaotao and others},
  journal={Nature},
  volume={619},
  pages={533--538},
  year={2023},
  publisher={Nature Publishing Group}
}

@misc{ncei2025billions,
  author={{NOAA National Centers for Environmental Information}},
  title={U.S. Billion-Dollar Weather and Climate Disasters},
  year={2025},
  doi={10.25921/stkw-7w73},
  url={https://www.ncei.noaa.gov/access/billions/}
}

@inproceedings{
ronneberger2015unet,
title={U-Net: Convolutional Networks for Biomedical Image Segmentation},
author="Ronneberger, Olaf
and Fischer, Philipp
and Brox, Thomas",
booktitle="Medical Image Computing and Computer-Assisted Intervention -- MICCAI 2015",
year="2015",
publisher="Springer International Publishing",
address="Cham",
pages="234--241",
isbn="978-3-319-24574-4"
}
\bibliographystyle{plain}

\appendix
\section{Additional Related Work}
\label{app:related}

\paragraph{Neural Process variants.}
Beyond the core NP family~\cite{garnelo2018conditional, garnelo2018neural}, cross-attention mechanisms improve context aggregation~\cite{kim2019anp}, spectral convolutions capture long-range dependencies in the frequency domain~\cite{mohseni2025specconvcnp}, and autoregressive extensions model non-Gaussian predictive distributions~\cite{bruinsma2023arcnp}.
NPs have been applied to climate downscaling~\cite{vaughan2022convnp_climate} and weather forecasting~\cite{allen2025end}, though the NP variants most relevant to this task still treat each time step independently.

\paragraph{Relationship to AR-CNP and FlowNP.}
AR-CNP~\cite{bruinsma2023arcnp} generates target predictions autoregressively, conditioning each new target point on previously predicted ones within a single time step.
This captures inter-target dependencies in the predictive distribution but does not introduce temporal dynamics across time steps---the axis addressed by NSP's transition loss $\mathcal{L}_{\mathrm{trans}}$.
Similarly, FlowNP~\cite{abuhamad2025flownp} replaces the Gaussian decoder with a conditional flow, improving single-timestep distributional quality without modeling temporal evolution.
Because both methods enhance \emph{within-timestep} prediction rather than \emph{across-timestep} coherence, the most informative NP comparison for our temporal contribution is ConvCNP~\cite{gordon2020convolutional}, which matches NSP's spatial convolutional structure while lacking any temporal component (Table~\ref{tab:performance_comparison_refined}).
We additionally note that both AR-CNP and FlowNP operate on explicit target-point sets; applying them directly to our $260 \times 590$ dense grid (${\sim}$153{,}000 target points per time step) would require sequential or chunked inference that substantially alters the original method and incurs prohibitive cost.

\paragraph{Neural SDE applications.}
Latent SDEs have been applied to irregular time series~\cite{oh2024stable, park2025acssm}, continuous-discrete state filtering~\cite{ansari2023neural}, and temporal point processes~\cite{zhang2024neural}.
Physics-informed Neural ODEs such as ClimODE~\cite{verma2024climode} incorporate advection principles for weather forecasting.
Simulation-free variants~\cite{park2025acssm, bartosh2025sde} avoid expensive trajectory-based training.

\section{Derivation of the Variational Objective}
\label{app:elbo}

We derive the evidence lower bound (ELBO) stated in Equation~(\ref{eq:elbo}) of the main text, then show that the transition KL admits a closed-form, simulation-free evaluation.

\subsection{ELBO decomposition}

Starting from the generative model in Equation~(\ref{eq:joint}), the marginal log-likelihood of the observations is
\begin{equation}
  \log p_\theta(\mathbf{y}_{0:T})
  = \log \int p_\theta(\mathbf{y}_{0:T}, \mathbf{z}_{0:T})\,d\mathbf{z}_{0:T}.
  \label{eq:app_marginal}
\end{equation}
Introducing the mean-field approximate posterior $q_\varphi(\mathbf{z}_{0:T} \mid \mathbf{x}_{0:T}) = \prod_{t=0}^{T} q_\varphi(\mathbf{z}_t \mid \mathbf{x}_t)$ and applying Jensen's inequality:
\begin{align}
  \log p_\theta(\mathbf{y}_{0:T})
  &= \log \int q_\varphi(\mathbf{z}_{0:T} \mid \mathbf{x}_{0:T})
     \frac{p_\theta(\mathbf{y}_{0:T}, \mathbf{z}_{0:T})}
          {q_\varphi(\mathbf{z}_{0:T} \mid \mathbf{x}_{0:T})}
     \,d\mathbf{z}_{0:T} \notag \\
  &\geq \mathbb{E}_{q_\varphi}\!\left[
    \log \frac{p_\theta(\mathbf{y}_{0:T}, \mathbf{z}_{0:T})}
              {q_\varphi(\mathbf{z}_{0:T} \mid \mathbf{x}_{0:T})}
  \right].
  \label{eq:app_jensen}
\end{align}
Substituting the factorizations of the generative model (\ref{eq:joint}) and the approximate posterior:
\begin{align}
  &\mathbb{E}_{q_\varphi}\!\left[
    \log \frac{p(\mathbf{z}_0)\prod_{t=0}^{T-1}p_\theta(\mathbf{z}_{t+1}\mid\mathbf{z}_t)\prod_{t=0}^{T}p_\theta(\mathbf{y}_t\mid\mathbf{z}_t)}
              {\prod_{t=0}^{T}q_\varphi(\mathbf{z}_t\mid\mathbf{x}_t)}
  \right] \notag \\
  &= \underbrace{\sum_{t=0}^{T}\mathbb{E}_{q_\varphi(\mathbf{z}_t\mid\mathbf{x}_t)}\!\bigl[\log p_\theta(\mathbf{y}_t\mid\mathbf{z}_t)\bigr]}_{-\mathcal{L}_{\mathrm{rec}}}
  \;-\; \underbrace{D_{\mathrm{KL}}\!\bigl(q_\varphi(\mathbf{z}_0\mid\mathbf{x}_0)\,\|\,p(\mathbf{z}_0)\bigr)}_{\mathcal{L}_{\mathrm{prior}}} \notag \\
  &\quad -\; \underbrace{\sum_{t=0}^{T-1}\mathbb{E}_{q_\varphi(\mathbf{z}_t\mid\mathbf{x}_t)}\!\Bigl[D_{\mathrm{KL}}\!\bigl(q_\varphi(\mathbf{z}_{t+1}\mid\mathbf{x}_{t+1})\,\|\,p_\theta(\mathbf{z}_{t+1}\mid\mathbf{z}_t)\bigr)\Bigr]}_{\mathcal{L}_{\mathrm{trans}}}.
  \label{eq:app_elbo_full}
\end{align}
The third line follows from grouping the $q_\varphi(\mathbf{z}_t \mid \mathbf{x}_t)$ terms for $t \geq 1$ with the corresponding transition $p_\theta(\mathbf{z}_t \mid \mathbf{z}_{t-1})$ and recognizing the resulting ratio as a KL divergence.
Rearranging gives the upper bound on the negative log-likelihood stated in Equation~(\ref{eq:elbo}).

The mean-field assumption $q_\varphi(\mathbf{z}_{0:T} \mid \mathbf{x}_{0:T}) = \prod_t q_\varphi(\mathbf{z}_t \mid \mathbf{x}_t)$ means the encoder processes each time step independently, with temporal coupling introduced solely through $\mathcal{L}_{\mathrm{trans}}$.
This avoids sequential inference at training time while still encouraging temporally consistent latent trajectories.

\subsection{Simulation-free transition KL}

Under the Euler--Maruyama discretization of the prior SDE (\ref{eq:prior_sde}) with step size $\Delta t$, the transition density is Gaussian (Equation~\ref{eq:euler}):
\begin{equation}
  p_\theta(\mathbf{z}_{t+1} \mid \mathbf{z}_t)
  = \mathcal{N}\!\bigl(
    \mathbf{z}_t + f_\theta(\mathbf{z}_t)\,\Delta t,\;
    \mathrm{diag}(\boldsymbol{\sigma}_\theta(\mathbf{z}_t)^2)\,\Delta t
  \bigr).
\end{equation}
Since both $q_\varphi(\mathbf{z}_{t+1} \mid \mathbf{x}_{t+1}) = \mathcal{N}(\boldsymbol{\mu}_\varphi(\mathbf{x}_{t+1}),\,\mathrm{diag}(\boldsymbol{\sigma}^2_\varphi(\mathbf{x}_{t+1})))$ and $p_\theta(\mathbf{z}_{t+1} \mid \mathbf{z}_t)$ are diagonal Gaussians, the KL divergence in $\mathcal{L}_{\mathrm{trans}}$ admits the closed form:
\begin{align}
  &D_{\mathrm{KL}}\!\bigl(q_\varphi(\mathbf{z}_{t+1}\mid\mathbf{x}_{t+1}) \,\|\, p_\theta(\mathbf{z}_{t+1}\mid\mathbf{z}_t)\bigr) \notag \\
  &= \frac{1}{2}\sum_{d=1}^{D'}\!\left[
    \log\frac{\sigma_{\theta,d}^2\,\Delta t}{\sigma_{\varphi,d}^2}
    + \frac{\sigma_{\varphi,d}^2 + (\mu_{\varphi,d} - z_{t,d} - f_{\theta,d}\,\Delta t)^2}
           {\sigma_{\theta,d}^2\,\Delta t}
    - 1
  \right],
  \label{eq:app_kl_gaussian}
\end{align}
where $D' = D \times H' \times W'$ is the total latent dimensionality, and subscript $d$ indexes individual elements.
This expression is computed in $\mathcal{O}(D')$ with no SDE solver calls, yielding the simulation-free training stated in the main text.

\subsection{Connection to Girsanov drift matching}

In the special case where the encoder variance matches the SDE diffusion, i.e.\ $\boldsymbol{\sigma}^2_\varphi(\mathbf{x}_{t+1}) = \boldsymbol{\sigma}_\theta(\mathbf{z}_t)^2\,\Delta t$, the log-ratio and variance terms in (\ref{eq:app_kl_gaussian}) cancel, and the KL reduces to
\begin{equation}
  \mathcal{L}_{\mathrm{trans}}
  = \frac{1}{2}\sum_{t=0}^{T-1}\mathbb{E}_{q_\varphi(\mathbf{z}_t\mid\mathbf{x}_t)}\!\left[
    \sum_{d=1}^{D'}\frac{(\mu_{\varphi,d}(\mathbf{x}_{t+1}) - z_{t,d} - f_{\theta,d}(\mathbf{z}_t)\,\Delta t)^2}
                        {\sigma_{\theta,d}(\mathbf{z}_t)^2\,\Delta t}
  \right].
  \label{eq:app_girsanov}
\end{equation}
This is equivalent to the Girsanov-based drift matching loss of Li et al.~\citep{li2020scalable}, where the numerator measures the discrepancy between the observed latent transition $\mu_{\varphi,d} - z_{t,d}$ and the predicted drift $f_{\theta,d}\,\Delta t$.
In the continuous-time limit ($\Delta t \to 0$), this recovers the path-space KL divergence between the posterior and prior SDE measures via Girsanov's theorem~\citep{li2020scalable, bartosh2025sde}.
Our formulation (\ref{eq:app_kl_gaussian}) generalizes this by allowing the encoder to learn an independent variance $\boldsymbol{\sigma}^2_\varphi$, providing additional flexibility for the approximate posterior.

\section{QPEBench Details}
\label{app:qpebench}

\subsection{Metric definitions}
\label{app:metrics}

$\text{RMSE}_g$ and $\text{MAE}_g$ quantify gauge-level error:
\begin{equation}
  \text{RMSE}_g = \sqrt{\frac{1}{N_g} \sum_{i=1}^{N_g} (\hat{y}_i - y^{(g)}_i)^2}, \quad
  \text{MAE}_g = \frac{1}{N_g} \sum_{i=1}^{N_g} |\hat{y}_i - y^{(g)}_i|,
\end{equation}
where $\hat{y}_i$ is the predicted precipitation, $y^{(g)}_i$ the gauge observation, and $N_g$ the number of gauge locations.
$\text{RMSE}_r$ and $\text{MAE}_r$ are defined identically with the radar reference $y^{(r)}_i$ and valid pixel count $N_r$.

$r_{r,\text{coll}}$ is the Pearson correlation at collocated precipitation points:
\begin{equation}
  r_{r,\text{coll}} = \frac{\sum_{i \in \mathcal{S}} (\hat{y}_i - \bar{\hat{y}}_{\mathcal{S}})(y^{(r)}_i - \bar{y}^{(r)}_{\mathcal{S}})}
  {\sqrt{\sum_{i \in \mathcal{S}} (\hat{y}_i - \bar{\hat{y}}_{\mathcal{S}})^2 \sum_{i \in \mathcal{S}} (y^{(r)}_i - \bar{y}^{(r)}_{\mathcal{S}})^2}},
\end{equation}
where $\mathcal{S} = \{i : y^{(g)}_i > \tau \;\text{and}\; y^{(r)}_i > \tau\}$ with $\tau = 0.1$\,mm/h.
Requiring both gauge and radar to exceed $\tau$ restricts evaluation to locations where two independent sensors agree that precipitation is occurring, eliminating correlation inflation by the dominant dry class and reducing representativeness error from point--area mismatch~\cite{tian2009component}.

$\text{FSS}_R$ is the Fractions Skill Score~\cite{roberts2008fss} averaged over four thresholds with a $20 \times 20$ pixel neighborhood ($2^\circ \times 2^\circ$):
\begin{equation}
  \text{FSS}(\tau) = 1 - \frac{\sum_{i}(O_i(\tau) - M_i(\tau))^2}{\sum_{i}O_i(\tau)^2 + \sum_{i}M_i(\tau)^2},
  \qquad
  \text{FSS}_R = \frac{1}{4}\sum_{\tau \in \Theta} \text{FSS}(\tau),
\end{equation}
where $O_i(\tau)$ and $M_i(\tau)$ are the observed and predicted fractions exceeding threshold $\tau$ within the neighborhood centered at pixel $i$, and $\Theta = \{1.0, 2.5, 5.0, 10.0\}$\,mm/h.

\begin{table}[!t]
  \centering
  \small
  \caption{QPEBench dataset overview.}
  \label{tab:dataset_overview}
  \setlength{\tabcolsep}{5pt}
  \renewcommand{\arraystretch}{1.1}
  \begin{tabular}{@{} l l @{}}
  \toprule
  \textbf{Property} & \textbf{Value} \\
  \midrule
  Temporal coverage   & January 2021 -- December 2025 (5 years) \\
  Total hourly steps  & 43{,}824 (43{,}756 after quality filtering) \\
  Grid resolution     & $0.1^\circ \times 0.1^\circ$ (${\sim}$11\,km) \\
  Grid size           & $260 \times 590$ (153{,}400 pixels) \\
  Spatial extent      & $24.05^\circ$--$49.95^\circ$N, $124.95^\circ$--$66.05^\circ$W \\
  \midrule
  Mean rainy fraction & 4.6\% $\pm$ 2.0\% of radar pixels per sample \\
  Mean gauge count    & 7{,}269 $\pm$ 382 stations per hour \\
  Gauge count range   & 4{,}312 -- 8{,}203 \\
  Gauge rain fraction & 5.4\% of reporting gauges \\
  \bottomrule
  \end{tabular}
  \end{table}

\begin{table}[!t]
\centering
\small
\caption{Distribution of non-zero radar precipitation across the full dataset.}
\label{tab:precip_dist}
\setlength{\tabcolsep}{8pt}
\renewcommand{\arraystretch}{1.1}
\begin{tabular}{@{} l r @{}}
\toprule
\textbf{Percentile} & \textbf{Value (mm/h)} \\
\midrule
P50  &   0.82 \\
P90  &   3.55 \\
P95  &   5.67 \\
P99  &  14.45 \\
Max  & 192.20 \\
\bottomrule
\end{tabular}
\end{table}

\subsection{Dataset statistics}
\label{app:dataset_stats}

Table~\ref{tab:dataset_overview} summarizes the QPEBench dataset used in all CONUS experiments.
The five-year period encompasses diverse meteorological conditions, including Hurricanes Ida (2021), Ian (2022), Helene (2024), and Milton (2024).
Missing rates are 0.00\% for GSMaP and 0.16\% for MRMS; the number of active gauges fluctuates hourly between 4{,}312 and 8{,}203 stations.

Table~\ref{tab:precip_dist} reports the distribution of non-zero radar precipitation, highlighting the extreme skewness typical of hourly rainfall data~\cite{trenberth2003changing}.
The zero-inflated, heavy-tailed nature of the distribution motivates the heteroscedastic Gaussian likelihood and the balanced context sampling strategy described in Appendix~\ref{app:hyperparams}.

\subsection{Quality filtering}
\label{app:filtering}

Four rules are applied before training:
(i)~\emph{gauge spike removal}: gauge values exceeding 20\,mm/h when both GSMaP and radar report ${<}\,1$\,mm/h are set to NaN;
(ii)~\emph{radar cap}: radar values exceeding 500\,mm/h are set to NaN;
(iii)~\emph{missing ground truth}: gauge values below $-900$ (sentinel) are excluded from the valid mask;
(iv)~\emph{completeness}: validation and test samples with entirely missing satellite or radar fields are discarded.

\subsection{Input normalization}
\label{app:normalization}

Table~\ref{tab:normalization} lists the transforms applied to each input channel inside the model's preprocessing step.
The radar target is kept in the original mm/h space; losses are computed without transformation.

\begin{table}[!t]
  \centering
  \small
  \caption{Input normalization.}
  \label{tab:normalization}
  \setlength{\tabcolsep}{5pt}
  \renewcommand{\arraystretch}{1.1}
  \begin{tabular}{@{} l l @{}}
  \toprule
  \textbf{Channel} & \textbf{Transform} \\
  \midrule
  Satellite (GSMaP)  & $\log(1 + x)$ \\
  Gauge observations & $\log(1 + x)$ \\
  Elevation (DEM)    & $x \,/\, 2000$ \\
  Context mask       & Binary $\{0,1\}$ \\
  Radar (target)     & None (raw mm/h) \\
  \bottomrule
  \end{tabular}
  \end{table}

\begin{table}[!t]
    \centering
    \caption{Hyperparameters of NSP.}
    \label{tab:setup}
    \begin{tabular}{ll}
    \toprule
    \textbf{Hyperparameter} & \textbf{Value} \\
    \midrule
    Optimizer & AdamW ($\beta_1{=}0.9$, $\beta_2{=}0.999$) \\
    Learning rate & $3 \times 10^{-3}$ (OneCycleLR) \\
    Weight decay & $1.35 \times 10^{-3}$ \\
    Epochs & 4 \\
    Batch size & 4 per GPU ($\times 8 = 32$ effective) \\
    Mixed precision & FP16 \\
    Gradient clipping & 1.0 \\
    \midrule
    Hidden channels & 128 \\
    Latent dimension $D$ & 64 \\
    Spatial downsampling ratio $r$ & 4 \\
    Encoder stages / residual blocks & 2 / 2 \\
    Decoder stages / residual blocks & 2 / 2 \\
    Decoder hidden dimension & 128 \\
    Fusion residual blocks & 3 \\
    SDE hidden dimension & 64 \\
    SDE step size $\Delta t$ & 1.0 \\
    Dropout & 0.1 \\
    Log-variance range & $[-6.0,\; {-}0.18]$ \\
    \midrule
    $\beta_{\mathrm{kl}}$ & 0.5 \\
    $\beta_{\mathrm{sde}}$ & 0.01 \\
    $\beta_{\mathrm{ctx}}$ & 15.0 \\
    $\beta_{\delta}$ & 90.0 \\
    Context ratio & 0.5 \\
    \bottomrule
    \end{tabular}
\end{table}

\section{Implementation Details}
\label{app:implementation}

\subsection{Hyperparameters}
\label{app:hyperparams}

\paragraph{Context/target protocol.}
At each training step, 50\% of gauge observations were randomly assigned to the context set $\mathcal{C}_t$ and the remainder to the target set $\mathcal{T}_t$, using balanced sampling that maintained equal proportions of rainy ($\geq 0.5$\,mm/h) and non-rainy observations in each partition.
The context set was bounded between 500 and 10{,}000 observations, and each sample required at least 16 rainy target locations.

\paragraph{Loss weights.}
The loss weights in the total objective were set to $\beta_{\mathrm{kl}}{=}0.5$, $\beta_{\mathrm{sde}}{=}0.01$, $\beta_{\mathrm{ctx}}{=}15$, and $\beta_{\delta}{=}90$.

\paragraph{Architecture and training.}
NSP contains 3.29M parameters in the encoder, 825K in the decoder, and 82K in the Neural SDE module (4.19M total).
Table~\ref{tab:setup} lists the full set of hyperparameters.

\subsection{Learned baseline training objective}
\label{app:baseline_loss}

For all learned baselines trained from sparse gauge supervision---ViT, U-Net, CNP, and ConvCNP---we used the same weighted held-out MSE objective.
The loss was evaluated only on held-out gauge targets, i.e.\ gauge locations not included in the context set.
This keeps the supervision consistent with the context-to-target protocol used throughout the paper and avoids rewarding models for reconstructing context gauges that were already provided as input.

Because held-out gauge targets are dominated by zero or weak-rain samples, a plain MSE objective tends to favor trivial low-rain solutions.
To mitigate this imbalance, we upweighted held-out rainy targets by a factor of 5.
Formally, for target set $\mathcal{T}$,
\begin{equation}
  \mathcal{L}_{\mathrm{base}}
  =
  \frac{\sum_{i \in \mathcal{T}} w_i \, (\hat{y}_i - y_i)^2}
       {\sum_{i \in \mathcal{T}} w_i},
  \qquad
  w_i =
  \begin{cases}
    5, & y_i \geq 0.5\ \mathrm{mm/h}, \\
    1, & y_i < 0.5\ \mathrm{mm/h}.
  \end{cases}
  \label{eq:baseline_weighted_mse}
\end{equation}
Here, $\hat{y}_i$ denotes the model prediction at held-out gauge location $i$, and $y_i$ is the corresponding gauge observation.
The target-weighting scheme was shared across all learned baselines for fairness.
Model-specific preprocessing, such as whether predictions were made in original or log space, followed the corresponding baseline configuration, but the supervision protocol itself---held-out targets only, with 5$\times$ weighting on rainy gauges above 0.5\,mm/h---was identical.

\subsection{ConvCNP implementation note}
\label{app:convcnp_impl}

Our ConvCNP baseline is not the original official code release, but a reimplementation adapted to dense-grid precipitation refinement.
The implementation preserves the key architectural idea of Gordon et al.~\cite{gordon2020convolutional}: sparse context observations are first mapped into a continuous representation by density-normalized Gaussian set convolution, and the resulting representation is then processed by a convolutional decoder to produce dense predictions.

Concretely, we applied a fixed Gaussian set convolution to the context gauge value map and context mask, yielding a smoothed value field and a smoothed density field.
These two fields were concatenated with the satellite precipitation and elevation channels, and the resulting four-channel tensor was passed through a U-Net-style encoder--decoder that outputs the predictive mean and log-variance on the full grid.
Thus, compared with the plain U-Net baseline, the distinguishing feature of ConvCNP is the convolutional deep-set preprocessing of sparse gauge observations rather than direct ingestion of the raw sparse context map.

This adaptation was necessary because the original ConvCNP formulation is not provided as an off-the-shelf implementation for our dense $260 \times 590$ gridded prediction setting.
Our reimplementation preserves the conditional process structure while adapting the decoder and training pipeline to dense-grid satellite precipitation refinement.

\begin{table}[!t]
\centering
\small
\caption{Expanding-window time-series cross-validation splits over the 2021--2025 period.}
\label{tab:cv_splits}
\begin{tabular}{@{} l l l l @{}}
\toprule
\textbf{Fold} & \textbf{Train} & \textbf{Validation} & \textbf{Test} \\
\midrule
1 & 2021 & 2022 & 2023 \\
2 & 2021--2022 & 2023 & 2024 \\
3 & 2021--2023 & 2024 & 2025 \\
\bottomrule
\end{tabular}
\end{table}

\begin{table}[!t]
\centering
\small
\caption{Full quantitative comparison on the Kyushu (Japan) regional dataset. Single train/validation/test split (2020/2021/2022); best in \textbf{bold}.}
\label{tab:kyushu_full}
\setlength{\tabcolsep}{5pt}
\renewcommand{\arraystretch}{1.15}
\begin{tabular}{@{} l ccccc @{}}
\toprule
\textbf{Method} & $\text{RMSE}_r \downarrow$ & $\text{MAE}_r \downarrow$ & $\text{RMSE}_g \downarrow$ & $\text{MAE}_g \downarrow$ & $r_{r,\text{coll}} \uparrow$ \\
\midrule
ViT                & 2.979 & 2.010 & 1.993 & 1.275 & 0.483 \\
Quantile mapping   & 1.876 & 1.073 & 1.377 & 0.623 & 0.216 \\
GWR                & 1.788 & 0.812 & 0.655 & 0.252 & 0.389 \\
CNP                & 1.779 & 1.559 & 1.953 & 1.815 & 0.437 \\
EMOS               & 1.738 & 0.954 & 1.082 & 0.326 & 0.169 \\
Cokriging          & 1.715 & 0.951 & 1.077 & 0.384 & 0.139 \\
Linear regression  & 1.688 & 0.831 & 0.768 & 0.275 & 0.356 \\
Kriging            & 1.640 & 0.936 & 0.973 & 0.364 & 0.114 \\
GSMaP              & 1.508 & 0.882 & 0.965 & 0.326 & 0.158 \\
XGBoost            & 1.501 & 0.835 & 0.985 & 0.404 & 0.190 \\
ConvCNP            & 1.363 & 0.752 & 0.645 & 0.170 & 0.525 \\
U-Net              & 1.324 & 0.726 & 0.679 & 0.186 & 0.532 \\
IDW                & 1.272 & 0.753 & 0.680 & 0.245 & 0.397 \\
\midrule
\rowcolor{gray!15}
GSMaP GC           & 1.288 & 0.808 & 0.762 & 0.270 & 0.265 \\
\rowcolor{highlightblue}
\textbf{NSP (Ours)} & \textbf{1.119} & \textbf{0.701} & \textbf{0.485} & \textbf{0.155} & \textbf{0.553} \\
\bottomrule
\end{tabular}
\end{table}

\subsection{Time-series cross-validation}
\label{app:cv}

Table~\ref{tab:cv_splits} details the three-fold expanding-window time-series cross-validation splits used for all CONUS experiments.
The training window grows with each fold, and the validation and test years always follow the training period chronologically, preventing temporal leakage.

\subsection{Temporal pairing}
\label{app:temporal}

For the SDE transition loss $\mathcal{L}_{\mathrm{trans}}$, consecutive hourly pairs $(t, t{+}1)$ are formed.
Pairs with a temporal gap exceeding one hour (due to quality filtering) are discarded.

\section{Additional Experimental Results}
\label{app:experiments}

\subsection{Kyushu regional experiment}
\label{app:kyushu}

\paragraph{Setup.}
The experiment covered the Kyushu region of Japan ($31.0$--$34.5^\circ$N, $129.0$--$132.5^\circ$E) on an approximately $80 \times 80$ grid at $0.1^\circ$ resolution.
Inputs consisted of GSMaP MVK satellite precipitation~\cite{kubota2020gsmap}, MERIT DEM elevation, and sparse AMeDAS rain gauge observations from approximately 100 stations; the Japan Meteorological Agency's RadarAMeDAS composite ($0.01^\circ$, regridded to $0.1^\circ$) served as the evaluation reference.
Data covered three calendar years (2020--2022).
Because only three years were available, we adopted a fixed year-based split (2020 training, 2021 validation, 2022 testing) instead of fold-based cross-validation.
The regional data are subject to institutional access restrictions.

\paragraph{Full results.}
Table~\ref{tab:kyushu_full} reports the full Kyushu results with all five metrics.
$\text{FSS}_R$ is omitted because the small grid size (${\sim}80 \times 80$) provides too few neighborhood patches for the statistic to be discriminative.

\FloatBarrier

\subsection{Extended error analysis}
\label{app:error}

This subsection provides extended error analysis for NSP on the CONUS test set (fold~3, 8{,}759 samples).

\begin{table}[!t]
\centering
\small
\caption{Minimum useful scale: smallest neighborhood where $\text{FSS} \geq 0.5$~\cite{roberts2008fss}. Lower indicates spatially sharper predictions.}
\label{tab:fss_scale}
\renewcommand{\arraystretch}{1.15}
\begin{tabular}{@{} l cc l @{}}
\toprule
\textbf{Threshold (mm/h)} & \textbf{GSMaP GC} & \textbf{NSP} & \textbf{Note} \\
\midrule
0.1  & 33\,km (3\,px)  & 33\,km (3\,px)  & Equal \\
1.0  & \textbf{33}\,km (3\,px)  & 55\,km (5\,px)  & GC better \\
5.0  & 440\,km (40\,px) & \textbf{220}\,km (20\,px) & NSP $2\times$ better \\
10.0 & not reached      & not reached      & Both below 0.5 \\
\bottomrule
\end{tabular}
\end{table}

\vspace{-2pt}
\paragraph{Minimum useful scale.}
Table~\ref{tab:fss_scale} reports the smallest neighborhood size at which FSS reaches 0.5, indicating the minimum scale at which the prediction becomes spatially useful.
At moderate intensity ($\tau{=}5.0$\,mm/h), NSP becomes useful at half the scale of GSMaP~GC (220\,km vs.\ 440\,km).
At light rain ($\tau{=}1.0$\,mm/h), GC's gauge-based daily adjustment provides slightly better spatial calibration.

\begin{table}[!t]
\centering
\small
\caption{Conditional intensity RMSE (cRMSE) and bias, computed only on pixels where both prediction and radar exceed the threshold.}
\label{tab:conditional_rmse}
\renewcommand{\arraystretch}{1.15}
\begin{tabular}{@{} l cc r rr @{}}
\toprule
\textbf{Threshold} & \textbf{GC} & \textbf{NSP} & \textbf{Improv.} & \textbf{NSP bias} & \textbf{GC bias} \\
\textbf{(mm/h)} & \textbf{cRMSE} & \textbf{cRMSE} & & & \\
\midrule
0.1  & 3.66  & \textbf{3.54}  & $+$3.3\%  & $-$0.52 & $-$0.34 \\
1.0  & 5.41  & \textbf{4.87}  & $+$10.0\% & $-$0.92 & $-$0.41 \\
5.0  & 9.72  & \textbf{9.24}  & $+$4.9\%  & $-$3.54 & $-$2.00 \\
10.0 & 13.38 & \textbf{11.88} & $+$11.2\% & $-$4.20 & $-$2.49 \\
25.0 & 21.32 & \textbf{14.70} & $+$31.1\% & $-$0.23 & $+$3.40 \\
\bottomrule
\end{tabular}
\end{table}

\vspace{-2pt}
\paragraph{Conditional intensity RMSE.}
Table~\ref{tab:conditional_rmse} reports RMSE computed only on pixels where both prediction and radar exceed the threshold, isolating intensity accuracy from detection accuracy.
NSP achieved lower conditional RMSE than GC at every threshold, with improvement growing from 3\% to 31\% as intensity increased.
Both methods underestimated precipitation (negative bias), but at $\tau{=}25$\,mm/h, GC overestimated ($+$3.40 bias) while NSP was near-zero ($-$0.23).

\paragraph{Why extreme precipitation remains difficult.}
The difficulty of extreme precipitation refinement stems from mutually reinforcing limitations at the input, resolution, training, and objective levels.

\emph{(i)~Input signal degradation.}
Microwave brightness temperatures saturate at high rain rates, and the retrieval assumptions underlying GSMaP---raindrop size distributions, vertical hydrometeor profiles, and surface emissivity models---break down during intense convection~\cite{kidd2011satellite, sun2018review_precip}.
Geolocation and parallax errors further displace convective cores by tens of kilometres in the satellite product~\cite{maggioni2016review_satellite_precip}.
Because NSP refines a residual on top of the satellite field, the model cannot recover structure that is absent or grossly misplaced in its input.

\emph{(ii)~Resolution and scale mismatch.}
Convective cells typically span 1--10\,km, while our grid resolution ($0.1^\circ \approx 11$\,km) can barely resolve them.
Both the radar reference and the satellite input represent area averages that smooth sub-grid peak intensities, imposing a ceiling on recoverable spatial detail.

\emph{(iii)~Training distribution imbalance.}
Extreme events are exceedingly rare: pixels exceeding 25\,mm/h constitute fewer than 0.01\% of all radar pixels (Table~\ref{tab:precip_dist}).
Standard minibatch sampling offers the model only a handful of extreme-intensity targets per epoch, providing insufficient gradient signal for the tail.

\emph{(iv)~Bayes-optimal smoothing under Gaussian likelihood.}
Under MSE-based training, the Bayes-optimal prediction converges toward the conditional mean, which systematically underestimates heavy-tailed extremes.
This is a mathematical property of the loss function, not a model deficiency: any method optimizing $L_2$ loss over a zero-inflated, heavy-tailed distribution will regress toward moderate values when the input is ambiguous.
The pixel-level RMSE breakdown confirms this: at 25+\,mm/h, all three methods (GSMaP, GC, NSP) converge to comparable RMSE (${\sim}$30\,mm/h), indicating that the information bottleneck lies in the input and the objective, not in model capacity.

\paragraph{Implications for gauge fusion.}
Sparse gauges can correct local bias where heavy rain is observed, but they cannot fully reconstruct a missed convective core when the satellite input already places precipitation in the wrong location or fails to detect it altogether.
This limitation is amplified by point--area mismatch: gauges observe point rainfall, whereas radar and satellite products represent area averages, so disagreement grows as spatial variability increases~\cite{tian2009component}.
Together with the failure case in Figure~\ref{fig:app_failure}, these results suggest that meaningful heavy-rain improvement will require advances along multiple axes---intensity-aware sampling or tail-weighted objectives to address the training imbalance, higher-resolution inputs or multi-scale architectures to resolve convective structure, and potentially distributional outputs (e.g., quantile regression) to avoid Bayes-optimal smoothing---rather than denser gauge conditioning alone.

\begin{table}[!t]
\centering
\small
\caption{MSE source decomposition into mutually exclusive pixel categories ($\tau{=}0.1$\,mm/h).}
\label{tab:mse_decomp}
\renewcommand{\arraystretch}{1.15}
\begin{tabular}{@{} ll r @{}}
\toprule
\textbf{Category} & \textbf{Definition} & \textbf{MSE share} \\
\midrule
Hit bias (intensity)  & pred $\geq 0.1$ and radar $\geq 0.1$ & 67.0\% \\
Miss (position)       & pred $< 0.1$ and radar $\geq 0.1$    & 14.4\% \\
False alarm (position) & pred $\geq 0.1$ and radar $< 0.1$   & 18.6\% \\
\midrule
\textbf{Intensity-related} & & \textbf{67.0\%} \\
\textbf{Position-related}  & & \textbf{33.0\%} \\
\bottomrule
\end{tabular}
\end{table}

\vspace{-2pt}
\paragraph{MSE source decomposition.}
Table~\ref{tab:mse_decomp} decomposes the total MSE into mutually exclusive pixel categories following GPM conventions.
Two-thirds of NSP's total error was attributable to intensity estimation at correctly detected locations; the remaining third arose from spatial errors (missed rain and false alarms).

\begin{table}[!t]
\centering
\small
\caption{Per-sample comparison conditioned on GSMaP GC accuracy (fold~3). 73~samples with missing GC data are excluded; 1~sample with GC RMSE${}\geq4$ is included in Overall but omitted from bins.}
\label{tab:gc_conditional}
\renewcommand{\arraystretch}{1.15}
\begin{tabular}{@{} l r cc r r @{}}
\toprule
\textbf{GC accuracy} & \textbf{$N$} & \textbf{GC RMSE} & \textbf{NSP RMSE} & \textbf{$\Delta$} & \textbf{NSP wins} \\
\midrule
RMSE $< 1$ (accurate) & 7{,}236 & 0.592 & 0.575 & $-$0.017 & 52.7\% \\
RMSE 1--2              & 1{,}415 & 1.193 & 1.014 & $-$0.179 & 81.0\% \\
RMSE 2--4              &      34 & 2.365 & 0.836 & $-$1.529 & 100.0\% \\
\midrule
\textbf{Overall}       & \textbf{8{,}686} & & & & \textbf{57.5\%} \\
\bottomrule
\end{tabular}
\end{table}

\vspace{-2pt}
\paragraph{GSMaP GC conditional analysis.}
Table~\ref{tab:gc_conditional} stratifies per-sample performance by GSMaP GC's accuracy.
When GC was already accurate (RMSE $< 1$), NSP matched it (52.7\% win rate, $\Delta{=}{-}0.017$).
When GC degraded (RMSE 1--2), NSP won 81\% of samples with a mean RMSE improvement of 0.179.

\begin{table}[!t]
\centering
\small
\caption{Per-sample comparison with GSMaP~GC and failure mode decomposition ($\tau{=}0.1$\,mm/h).}
\label{tab:failure_decomp}
\setlength{\tabcolsep}{5pt}
\renewcommand{\arraystretch}{1.15}
\begin{tabular}{@{} l rr @{}}
\toprule
& \textbf{Count} & \textbf{Proportion} \\
\midrule
NSP better than GC & 4{,}994 & 57.5\% \\
NSP worse than GC  & 3{,}692 & 42.5\% \\
\midrule
\multicolumn{3}{@{}l}{\emph{Dominant error in 3{,}692 failure cases}} \\
\quad Intensity error (hit bias) & 2{,}681 & 72.6\% \\
\quad False precipitation        &    571 & 15.5\% \\
\quad Missed precipitation       &    440 & 11.9\% \\
\bottomrule
\end{tabular}
\end{table}

\vspace{-2pt}
\paragraph{Per-sample failure decomposition.}
Table~\ref{tab:failure_decomp} classifies the 8{,}686 test samples for which GC data are available by their dominant error source following the GPM hit-bias/miss/false-alarm decomposition ($\tau{=}0.1$\,mm/h).
NSP outperformed GSMaP~GC in 57.5\% of hourly samples.
Among the 3{,}692 failure cases, 72.6\% were dominated by intensity error at correctly detected locations, while spatial errors (false alarms and misses) accounted for the remaining 27.4\%.
Pixel-level MSE decomposition confirmed that 67\% of total error is intensity-related.

\begin{table}[!t]
\centering
\small
\caption{Dominant error mode for all 8{,}759 test samples ($\tau{=}0.1$\,mm/h).}
\label{tab:failure_all}
\renewcommand{\arraystretch}{1.15}
\begin{tabular}{@{} l rr @{}}
\toprule
\textbf{Mode} & \textbf{Count} & \textbf{Proportion} \\
\midrule
Intensity error (hit bias) & 7{,}125 & 81.3\% \\
False precipitation        &    930 & 10.6\% \\
Missed precipitation       &    704 &  8.0\% \\
\bottomrule
\end{tabular}
\end{table}

\paragraph{Failure mode across all samples.}
Table~\ref{tab:failure_all} shows the dominant error mode for all test samples.
Intensity error dominated in 81\% of cases; false alarms (11\%) and misses (8\%) were secondary contributors.

\FloatBarrier

\subsection{Theory-motivated temporal ablations}
\label{app:temporal_ablation}

\begin{table}[!t]
\centering
\small
\caption{Temporal objective variants on CONUS. The matched-variance row uses the Girsanov special case $\sigma_\varphi^2 = \sigma_\theta^2 \Delta t$; the naive temporal penalty replaces the KL with mean-only temporal matching. Best values are in \textbf{bold}.}
\label{tab:temporal_ablation}
\setlength{\tabcolsep}{4.5pt}
\renewcommand{\arraystretch}{1.1}
\begin{tabular}{@{} l cccccc @{}}
\toprule
\textbf{Variant} & $\text{RMSE}_r \downarrow$ & $\text{MAE}_r \downarrow$ & $\text{RMSE}_g \downarrow$ & $\text{MAE}_g \downarrow$ & $r_{r,\text{coll}} \uparrow$ & $\text{FSS}_R \uparrow$ \\
\midrule
Matched variance & 2.883{\scriptsize $\pm$0.139} & 1.489{\scriptsize $\pm$0.076} & \textbf{0.341}{\scriptsize $\pm$0.062} & \textbf{0.063}{\scriptsize $\pm$0.015} & \textbf{0.478}{\scriptsize $\pm$0.021} & 0.522{\scriptsize $\pm$0.018} \\
Naive temporal penalty & 2.853{\scriptsize $\pm$0.121} & 1.470{\scriptsize $\pm$0.089} & 0.366{\scriptsize $\pm$0.037} & 0.070{\scriptsize $\pm$0.011} & \textbf{0.478}{\scriptsize $\pm$0.018} & 0.519{\scriptsize $\pm$0.042} \\
\rowcolor{highlightblue}
Full transition KL & \textbf{2.818}{\scriptsize $\pm$0.062} & \textbf{1.444}{\scriptsize $\pm$0.026} & 0.393{\scriptsize $\pm$0.047} & 0.076{\scriptsize $\pm$0.013} & \textbf{0.478}{\scriptsize $\pm$0.021} & \textbf{0.527}{\scriptsize $\pm$0.022} \\
\bottomrule
\end{tabular}
\end{table}

\begin{figure}[!t]
\centering
\begin{tikzpicture}
\begin{groupplot}[
  group style={
    group size=2 by 2,
    horizontal sep=1.6cm,
    vertical sep=1.6cm
  },
  width=0.44\textwidth,
  height=4.6cm,
  xlabel={Hour},
  xmin=-0.2, xmax=11.2,
  xtick={0,2,4,6,8,10},
  grid=major, grid style={gray!30},
  tick label style={font=\small},
  label style={font=\small},
  title style={font=\small, yshift=-0.5ex},
  legend style={font=\scriptsize, draw=none, fill=white, fill opacity=0.8},
]
\nextgroupplot[
  ylabel={$\mathrm{RMSE}_r$},
  ymin=1.8, ymax=3.7,
  title={(a) March 15, 2025: $\mathrm{RMSE}_r$},
  legend to name=eventtrackinglegend,
  legend columns=3,
  legend cell align=left,
]
\addplot[blue, thick, mark=*, mark size=1.9] coordinates {(0,1.9439)(1,2.1722)(2,2.5510)(3,2.8864)(4,3.3572)(5,3.2988)(6,2.6714)(7,3.0581)(8,2.8706)(9,2.9092)(10,2.7588)(11,2.7591)};
\addlegendentry{NSP}
\addplot[orange, thick, mark=square*, mark size=1.9] coordinates {(0,1.9577)(1,2.2223)(2,2.6756)(3,2.9772)(4,3.4269)(5,3.3790)(6,2.7541)(7,3.0468)(8,2.9919)(9,2.9646)(10,2.8230)(11,2.7125)};
\addlegendentry{NSP w/o $\mathcal{L}_{\mathrm{trans}}$}
\addplot[teal, thick, mark=triangle*, mark size=2.2] coordinates {(0,2.1442)(1,2.4143)(2,2.8963)(3,3.3155)(4,3.6883)(5,3.5433)(6,3.0665)(7,3.2731)(8,3.3246)(9,3.2931)(10,3.3255)(11,3.1892)};
\addlegendentry{ConvCNP}

\nextgroupplot[
  ylabel={$\mathrm{FSS}_R$},
  ymin=0.0, ymax=0.9,
  title={(b) March 15, 2025: $\mathrm{FSS}_R$},
]
\addplot[blue, thick, mark=*, mark size=1.9] coordinates {(0,0.5644)(1,0.6902)(2,0.7539)(3,0.8239)(4,0.8465)(5,0.7704)(6,0.6735)(7,0.6727)(8,0.6813)(9,0.6252)(10,0.7101)(11,0.6669)};
\addplot[orange, thick, mark=square*, mark size=1.9] coordinates {(0,0.5702)(1,0.6457)(2,0.7090)(3,0.8336)(4,0.8040)(5,0.7612)(6,0.6716)(7,0.5906)(8,0.5900)(9,0.6172)(10,0.6518)(11,0.6012)};
\addplot[teal, thick, mark=triangle*, mark size=2.2] coordinates {(0,0.0380)(1,0.0509)(2,0.0883)(3,0.1163)(4,0.1000)(5,0.1086)(6,0.0742)(7,0.0427)(8,0.0585)(9,0.0534)(10,0.0487)(11,0.0532)};

\nextgroupplot[
  ylabel={$\mathrm{RMSE}_r$},
  ymin=1.5, ymax=3.0,
  title={(c) October 30--31, 2025: $\mathrm{RMSE}_r$},
]
\addplot[blue, thick, mark=*, mark size=1.9] coordinates {(0,1.9902)(1,2.0114)(2,2.1759)(3,2.1543)(4,1.8766)(5,1.8745)(6,1.8136)(7,2.0268)(8,1.9405)(9,1.6416)(10,1.6573)(11,1.5435)};
\addplot[orange, thick, mark=square*, mark size=1.9] coordinates {(0,2.1163)(1,2.4198)(2,2.8594)(3,2.8750)(4,2.2426)(5,2.3388)(6,2.1369)(7,2.4475)(8,2.3785)(9,2.0359)(10,1.7448)(11,1.9474)};
\addplot[teal, thick, mark=triangle*, mark size=2.2] coordinates {(0,2.4093)(1,2.4049)(2,2.4285)(3,2.3548)(4,2.2444)(5,2.2442)(6,2.3113)(7,2.2702)(8,1.9777)(9,1.7757)(10,1.8461)(11,1.7303)};

\nextgroupplot[
  ylabel={$\mathrm{FSS}_R$},
  ymin=0.0, ymax=0.8,
  title={(d) October 30--31, 2025: $\mathrm{FSS}_R$},
]
\addplot[blue, thick, mark=*, mark size=1.9] coordinates {(0,0.5956)(1,0.4970)(2,0.5352)(3,0.5554)(4,0.5654)(5,0.5711)(6,0.5845)(7,0.4468)(8,0.5311)(9,0.7449)(10,0.4334)(11,0.4116)};
\addplot[orange, thick, mark=square*, mark size=1.9] coordinates {(0,0.5887)(1,0.4622)(2,0.4583)(3,0.4467)(4,0.4799)(5,0.4567)(6,0.4912)(7,0.4108)(8,0.4424)(9,0.5922)(10,0.4938)(11,0.3466)};
\addplot[teal, thick, mark=triangle*, mark size=2.2] coordinates {(0,0.0408)(1,0.0344)(2,0.0511)(3,0.0504)(4,0.0240)(5,0.0221)(6,0.0321)(7,0.0397)(8,0.0350)(9,0.0486)(10,0.0215)(11,0.0187)};
\end{groupplot}
\node[anchor=north, yshift=-10mm] at ($(group c1r2.south)!0.5!(group c2r2.south)$) {\pgfplotslegendfromname{eventtrackinglegend}};
\end{tikzpicture}
\caption{Hour-by-hour event tracking using the existing radar-evaluated metrics $\mathrm{RMSE}_r \downarrow$ and $\mathrm{FSS}_R \uparrow$ on two 12-hour fold~3 events. NSP is compared against the no-$\mathcal{L}_{\mathrm{trans}}$ ablation and ConvCNP. In both events, removing $\mathcal{L}_{\mathrm{trans}}$ degrades $\mathrm{RMSE}_r$ and usually lowers $\mathrm{FSS}_R$, with the effect especially pronounced in the late-October event. ConvCNP improves over naive gauge-only interpolation but still fails to preserve thresholded spatial structure, yielding consistently low $\mathrm{FSS}_R$ under dense-grid prediction.}
\label{fig:event_tracking}
\end{figure}
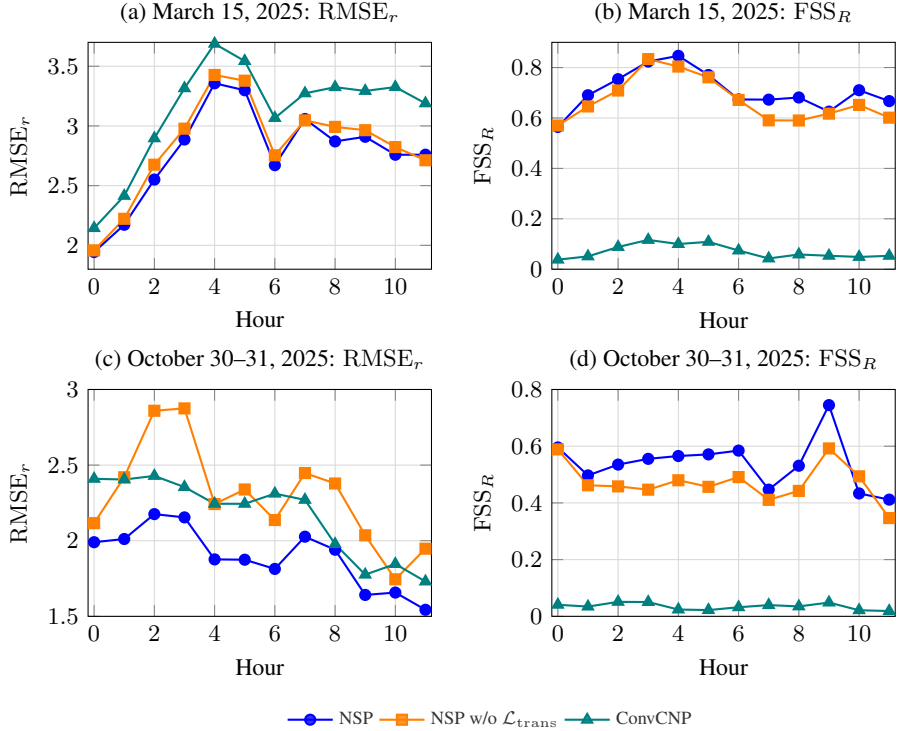

Table~\ref{tab:temporal_ablation} compares the proposed transition KL with two theory-motivated variants: a matched-variance special case corresponding to the Girsanov reduction, and a naive temporal penalty that replaces the KL with mean-only temporal matching.
The matched-variance variant achieves the lowest gauge-level errors, but the full transition KL yields the best radar-evaluated gridded accuracy ($\text{RMSE}_r$: 2.818, $\text{MAE}_r$: 1.444) and the highest spatial coherence ($\text{FSS}_R$: 0.527).
The naive temporal penalty is competitive but remains weaker than the full KL on all three gridded metrics.
These results support the use of the full transition KL as the best trade-off between pointwise gauge fitting and spatially coherent gridded prediction.
The matched-variance constraint ties the encoder's uncertainty budget to the SDE diffusion, which reduces the posterior's capacity to absorb observation noise independently of the dynamics; the full KL avoids this coupling and lets the encoder and the SDE each adapt their variance scales, yielding better gridded accuracy at a modest cost in gauge-level error.

\FloatBarrier

\subsection{Event-wise temporal analysis}
\label{app:event_temporal}

To probe the effect of temporal regularization without introducing additional evaluation metrics, we tracked the existing event-level metrics $\mathrm{RMSE}_r$ and $\mathrm{FSS}_R$ hour by hour over two 12-hour CONUS events in the 2025 test split (fold~3): a rapidly evolving severe-weather event on March~15, 2025, and a late-October synoptic-scale precipitation event on October~30--31, 2025.
We compared NSP against the no-$\mathcal{L}_{\mathrm{trans}}$ ablation and ConvCNP, the latter implemented as a faithful dense-grid reimplementation described in Appendix~\ref{app:convcnp_impl} and trained with the shared weighted held-out MSE objective of Appendix~\ref{app:baseline_loss}.

Figure~\ref{fig:event_tracking} shows the hour-by-hour trajectories, while Table~\ref{tab:event_temporal} summarizes the event-wise means.
Across the two events, NSP outperformed the no-$\mathcal{L}_{\mathrm{trans}}$ ablation in 22/24 hourly steps in $\mathrm{RMSE}_r$ and 21/24 hourly steps in $\mathrm{FSS}_R$.
The gain was especially pronounced in the late-October event, where removing $\mathcal{L}_{\mathrm{trans}}$ increased the event-averaged $\mathrm{RMSE}_r$ from 1.892 to 2.295 and reduced $\mathrm{FSS}_R$ from 0.539 to 0.472, while also increasing the hour-to-hour standard deviation of $\mathrm{RMSE}_r$ from 0.191 to 0.323.
This indicates that the transition loss improves stability over multi-hour evolution rather than only average test-set performance.

\begin{table}[!t]
\centering
\small
\caption{Event-wise temporal analysis on two 12-hour CONUS events in fold~3. Values are means over the 12 hourly steps; $\sigma(\mathrm{RMSE}_r)$ denotes the standard deviation of $\mathrm{RMSE}_r$ across the event window. ConvCNP denotes a reimplementation adapted to dense-grid prediction (Appendix~\ref{app:convcnp_impl}).}
\label{tab:event_temporal}
\setlength{\tabcolsep}{6pt}
\renewcommand{\arraystretch}{1.12}
\begin{tabular}{@{} l l c c c @{}}
\toprule
\textbf{Event} & \textbf{Model} & $\mathrm{RMSE}_r \downarrow$ & $\sigma(\mathrm{RMSE}_r) \downarrow$ & $\mathrm{FSS}_R \uparrow$ \\
\midrule
March 15, 2025 & NSP & \textbf{2.770} & \textbf{0.392} & \textbf{0.707} \\
March 15, 2025 & NSP w/o $\mathcal{L}_{\mathrm{trans}}$ & 2.828 & 0.403 & 0.671 \\
March 15, 2025 & ConvCNP & 3.123 & 0.427 & 0.069 \\
\midrule
October 30--31, 2025 & NSP & \textbf{1.892} & \textbf{0.191} & \textbf{0.539} \\
October 30--31, 2025 & NSP w/o $\mathcal{L}_{\mathrm{trans}}$ & 2.295 & 0.323 & 0.472 \\
October 30--31, 2025 & ConvCNP & 2.166 & 0.249 & 0.035 \\
\bottomrule
\end{tabular}
\end{table}

\begin{table}[!t]
\centering
\small
\caption{SDE rollout analysis on the same two 12-hour fold~3 events. NSP (standard inference) denotes the deployed inference mode used throughout the paper, which re-encodes each hour independently. NSP (pure SDE rollout) propagates the latent state without re-encoding after initialization, and NSP (hybrid) linearly blends rolled-out and re-encoded latent states with $\alpha=0.5$. Values are means over the 12 hourly steps; latency is the mean per-hour inference time within each mode.}
\vspace{1mm}
\label{tab:rollout_summary}
\setlength{\tabcolsep}{7pt}
\renewcommand{\arraystretch}{1.12}
\begin{tabular}{@{} l l c c c @{}}
\toprule
\textbf{Event} & \textbf{Mode} & $\mathrm{RMSE}_r \downarrow$ & $\mathrm{FSS}_R \uparrow$ & \textbf{Mean latency (ms)} \\
\midrule
March 15, 2025 & NSP (standard inference) & \textbf{2.776} & \textbf{0.713} & 2.60 \\
March 15, 2025 & NSP (pure SDE rollout) & 3.239 & 0.351 & \textbf{2.22} \\
March 15, 2025 & NSP (hybrid, $\alpha{=}0.5$) & 3.102 & 0.497 & 2.92 \\
\midrule
October 30--31, 2025 & NSP (standard inference) & 1.891 & \textbf{0.543} & 2.60 \\
October 30--31, 2025 & NSP (pure SDE rollout) & 2.043 & 0.313 & \textbf{2.21} \\
October 30--31, 2025 & NSP (hybrid, $\alpha{=}0.5$) & \textbf{1.851} & 0.503 & 2.93 \\
\bottomrule
\end{tabular}
\end{table}

ConvCNP performed markedly worse on both events.
In our dense-grid setting, the ConvCNP baseline converged to smooth low-intensity fields that rarely exceeded the spatial verification thresholds, yielding very low event-averaged $\mathrm{FSS}_R$ (0.069 on March~15 and 0.035 on October~30--31).
This gap persists despite ConvCNP sharing a similar convolutional backbone and producing heteroscedastic output; the critical difference is that ConvCNP lacks a latent variable and instead relies on a deterministic set-convolution encoder, which collapses all context information into a single smoothed representation before decoding.
NSP's stochastic latent field allows the decoder to sample from a distribution that can represent spatially varying modes---rain versus no-rain---at each location, whereas the deterministic ConvCNP pathway averages over this ambiguity.
The result confirms that a convolutional spatial encoder alone is insufficient to preserve thresholded spatial structure under dense gridded inference, and that the stochastic latent pathway of NSP is essential.

\FloatBarrier

\subsection{SDE rollout analysis}
\label{app:sde_rollout}

To assess whether the learned SDE captures meaningful latent dynamics beyond its role as a training objective, we compared three inference modes on the same two 12-hour events: (i) standard NSP inference with per-hour re-encoding, (ii) pure SDE rollout after the initial latent state, and (iii) a hybrid mode that linearly blends re-encoded and rolled-out latent states.

Table~\ref{tab:rollout_summary} summarizes the results.
NSP (standard inference) is the deployed inference mode used throughout the paper and is slower than pure rollout because it re-encodes each hour, whereas pure rollout skips those encoder passes after initialization.
Pure rollout remained competitive for the first few hours and retained $\mathrm{FSS}_R > 0.3$ for the first 6 hours in the March event and 5 hours in the late-October event, confirming that the learned SDE captures non-trivial short-horizon dynamics.
Beyond that horizon, rollout quality degraded rapidly, consistent with accumulated latent drift.
The hybrid mode was more stable than pure rollout and, in the late-October event, even achieved lower mean $\mathrm{RMSE}_r$ than NSP (standard inference) (1.851 vs.\ 1.891) while retaining $\mathrm{FSS}_R > 0.3$ for 10 of 12 hours, suggesting that adaptive blending may be a promising direction for future work.
These observations are consistent with the mean-field posterior assumption (Section~4.3): because the encoder produces each $\mathbf{z}_t$ independently, the SDE's drift $f_\theta$ can only approximate the marginal transition between independently inferred latent states.
Over short horizons the approximation is accurate, but errors compound because no new observational information corrects the latent trajectory.
Re-encoding at each hour resets this drift and reintroduces the current observation, which is why standard inference remains the most robust strategy over longer horizons.
The hybrid result nonetheless suggests that partially incorporating the SDE at inference time can recover complementary temporal information that the encoder alone does not capture, pointing toward adaptive encoder--rollout blending as a direction for future work.

\FloatBarrier

\subsection{Gauge density sensitivity analysis}
\label{app:gauge_sensitivity}

\begin{figure}[!t]
\centering
\begin{tikzpicture}
\begin{axis}[
  name=left,
  width=0.48\textwidth, height=5.2cm,
  xlabel={Context gauge ratio (\%)},
  ylabel={Metric value},
  xmin=5, xmax=105,
  xtick={10,25,50,75,100},
  legend style={at={(0.98,0.50)}, anchor=east, font=\scriptsize, draw=none, fill=white, fill opacity=0.8},
  grid=major, grid style={gray!30},
  title={\small (a) Radar-evaluated metrics},
  tick label style={font=\small},
  label style={font=\small},
]
\addplot[blue, thick, mark=*, mark size=2] coordinates {(10,2.733)(25,2.735)(50,2.748)(75,2.763)(100,2.782)};
\addlegendentry{$\text{RMSE}_r$ ($\downarrow$)}
\addplot[red, thick, mark=square*, mark size=2] coordinates {(10,0.490)(25,0.501)(50,0.490)(75,0.475)(100,0.460)};
\addlegendentry{$\text{FSS}_R$ ($\uparrow$)}
\addplot[teal, thick, mark=triangle*, mark size=2.5] coordinates {(10,0.449)(25,0.460)(50,0.454)(75,0.448)(100,0.441)};
\addlegendentry{$r_{r,\text{coll}}$ ($\uparrow$)}
\end{axis}
\begin{axis}[
  at={(left.east)}, anchor=west, xshift=12mm,
  width=0.48\textwidth, height=5.2cm,
  xlabel={Context gauge ratio (\%)},
  ylabel={Metric value},
  xmin=5, xmax=105,
  xtick={10,25,50,75,100},
  legend style={at={(0.98,0.98)}, anchor=north east, font=\scriptsize, draw=none, fill=white, fill opacity=0.8},
  grid=major, grid style={gray!30},
  title={\small (b) Gauge-evaluated metrics},
  tick label style={font=\small},
  label style={font=\small},
]
\addplot[blue, thick, mark=*, mark size=2] coordinates {(10,0.568)(25,0.476)(50,0.425)(75,0.398)(100,0.385)};
\addlegendentry{$\text{RMSE}_g$ ($\downarrow$)}
\addplot[red, thick, mark=square*, mark size=2] coordinates {(10,0.134)(25,0.108)(50,0.084)(75,0.071)(100,0.064)};
\addlegendentry{$\text{MAE}_g$ ($\downarrow$)}
\end{axis}
\end{tikzpicture}
\caption{Effect of the inference-time context ratio on NSP performance (fold~3, model trained with 50\% context). (a)~Radar-evaluated metrics are best at 10--25\% context and degrade with denser input, revealing a gauge over-interpolation effect. (b)~Gauge-evaluated metrics improve monotonically with denser context, as expected.}
\label{fig:gauge_sensitivity}
\end{figure}
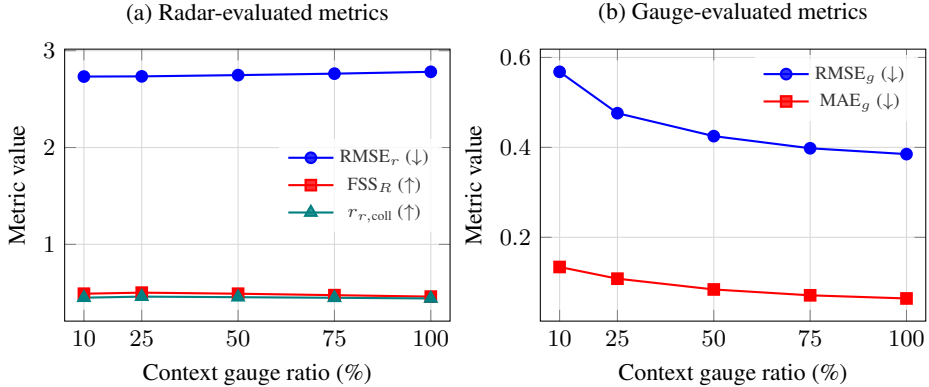

\paragraph{Test-time context ratio.}
We varied the fraction of available gauges assigned to the context set at test time from 10\% to 100\%, keeping the model fixed (fold~3, trained with 50\% context).
Figure~\ref{fig:gauge_sensitivity} plots the results.
Radar-evaluated metrics peak at 10--25\% context and degrade with denser input: going from 10\% to 100\% worsens $\text{RMSE}_r$ by 1.8\% and $\text{FSS}_R$ by 6.0\%, consistent with the decoder over-interpolating toward individual gauge values at the expense of spatial coherence (Section~\ref{sec:results}).
Gauge-evaluated metrics improve monotonically, as expected when context gauges overlap with the evaluation set.
The most balanced operating point is 25\% context ($\text{FSS}_R{=}0.501$, $r_{r,\text{coll}}{=}0.460$).
Even at 10\% (${\sim}$730 stations, ${\sim}$100\,km spacing), NSP retains spatial accuracy on par with 100\%, confirming robustness to sparse observation networks.

\begin{figure}[!t]
  \centering
  \begin{tikzpicture}
  \begin{axis}[
    name=left,
    width=0.48\textwidth, height=5.2cm,
    xlabel={Training context ratio (\%)},
    ylabel={Metric value},
    xmin=5, xmax=80,
    xtick={10,25,50,75},
    legend style={at={(0.98,0.50)}, anchor=east, font=\scriptsize, draw=none, fill=white, fill opacity=0.8},
    grid=major, grid style={gray!30},
    title={\small (a) Radar-evaluated metrics},
    tick label style={font=\small},
    label style={font=\small},
  ]
  \addplot[blue, thick, mark=*, mark size=2] coordinates {(10,2.728)(25,2.746)(50,2.754)(75,2.693)};
  \addlegendentry{$\text{RMSE}_r$ ($\downarrow$)}
  \addplot[red, thick, mark=square*, mark size=2] coordinates {(10,0.468)(25,0.508)(50,0.508)(75,0.335)};
  \addlegendentry{$\text{FSS}_R$ ($\uparrow$)}
  \addplot[teal, thick, mark=triangle*, mark size=2.5] coordinates {(10,0.444)(25,0.457)(50,0.449)(75,0.458)};
  \addlegendentry{$r_{r,\text{coll}}$ ($\uparrow$)}
  \end{axis}
  \begin{axis}[
    at={(left.east)}, anchor=west, xshift=12mm,
    width=0.48\textwidth, height=5.2cm,
    xlabel={Training context ratio (\%)},
    ylabel={Metric value},
    xmin=5, xmax=80,
    xtick={10,25,50,75},
    legend style={at={(0.98,0.98)}, anchor=north east, font=\scriptsize, draw=none, fill=white, fill opacity=0.8},
    grid=major, grid style={gray!30},
    title={\small (b) Gauge-evaluated metrics},
    tick label style={font=\small},
    label style={font=\small},
  ]
  \addplot[blue, thick, mark=*, mark size=2] coordinates {(10,0.574)(25,0.427)(50,0.420)(75,0.825)};
  \addlegendentry{$\text{RMSE}_g$ ($\downarrow$)}
  \addplot[red, thick, mark=square*, mark size=2] coordinates {(10,0.167)(25,0.089)(50,0.083)(75,0.447)};
  \addlegendentry{$\text{MAE}_g$ ($\downarrow$)}
  \end{axis}
  \end{tikzpicture}
  \caption{Effect of the training-time context ratio on NSP performance (fold~3, seed 42). Each point retrains NSP from scratch with the indicated fixed context ratio; the 100\% setting diverged and is omitted. At 75\%, $\text{FSS}_R$ collapses despite a slight $\text{RMSE}_r$ improvement, confirming that dense gauge supervision during training induces overfitting to point observations.}
  \label{fig:train_density}
  \end{figure}
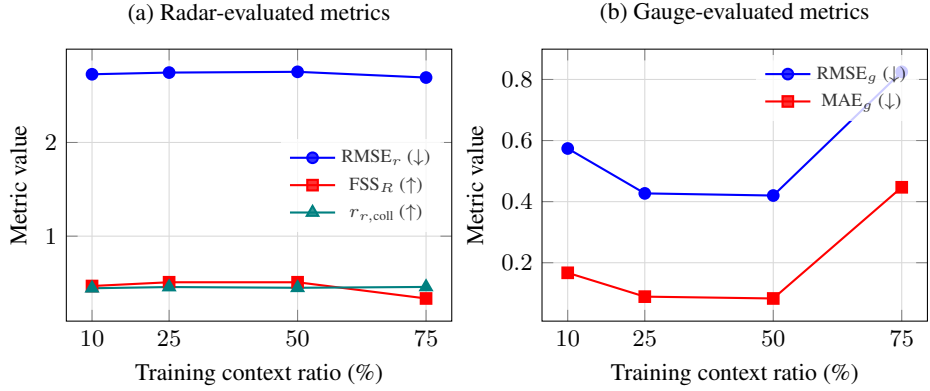

\paragraph{Training-time context ratio.}
To disentangle test-time over-interpolation from the training protocol, we retrained NSP from scratch with a fixed context ratio.
Figure~\ref{fig:train_density} reports the results.
The 100\% setting diverged; 25--50\% yielded the best balance between gridded and gauge-level accuracy.
At 75\%, $\text{RMSE}_r$ fell to 2.693 but $\text{FSS}_R$ dropped to 0.335 and $\text{RMSE}_g$ rose to 0.825, confirming that denser gauge supervision during training also induces overfitting to point observations.

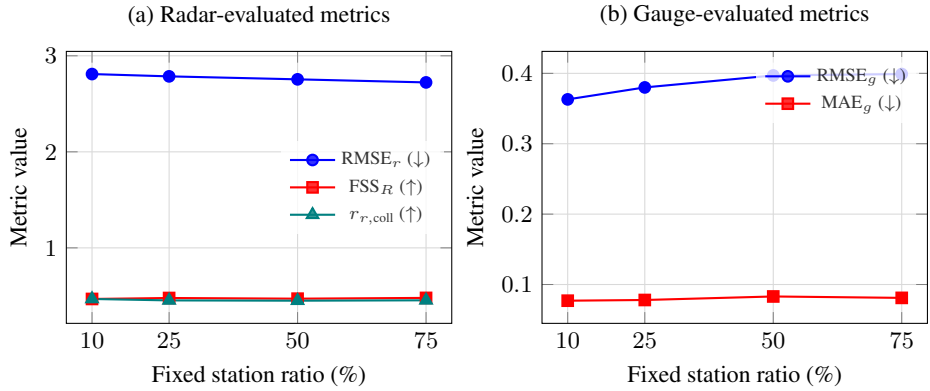
\begin{figure}[!t]
\centering
\begin{tikzpicture}
\begin{axis}[
  name=left,
  width=0.48\textwidth, height=5.2cm,
  xlabel={Fixed station ratio (\%)},
  ylabel={Metric value},
  xmin=5, xmax=80,
  xtick={10,25,50,75},
  legend style={at={(0.98,0.50)}, anchor=east, font=\scriptsize, draw=none, fill=white, fill opacity=0.8},
  grid=major, grid style={gray!30},
  title={\small (a) Radar-evaluated metrics},
  tick label style={font=\small},
  label style={font=\small},
]
\addplot[blue, thick, mark=*, mark size=2] coordinates {(10,2.810)(25,2.786)(50,2.755)(75,2.723)};
\addlegendentry{$\text{RMSE}_r$ ($\downarrow$)}
\addplot[red, thick, mark=square*, mark size=2] coordinates {(10,0.468)(25,0.478)(50,0.471)(75,0.479)};
\addlegendentry{$\text{FSS}_R$ ($\uparrow$)}
\addplot[teal, thick, mark=triangle*, mark size=2.5] coordinates {(10,0.468)(25,0.453)(50,0.450)(75,0.454)};
\addlegendentry{$r_{r,\text{coll}}$ ($\uparrow$)}
\end{axis}
\begin{axis}[
  at={(left.east)}, anchor=west, xshift=12mm,
  width=0.48\textwidth, height=5.2cm,
  xlabel={Fixed station ratio (\%)},
  ylabel={Metric value},
  xmin=5, xmax=80,
  xtick={10,25,50,75},
  legend style={at={(0.98,0.98)}, anchor=north east, font=\scriptsize, draw=none, fill=white, fill opacity=0.8},
  grid=major, grid style={gray!30},
  title={\small (b) Gauge-evaluated metrics},
  tick label style={font=\small},
  label style={font=\small},
]
\addplot[blue, thick, mark=*, mark size=2] coordinates {(10,0.363)(25,0.380)(50,0.397)(75,0.399)};
\addlegendentry{$\text{RMSE}_g$ ($\downarrow$)}
\addplot[red, thick, mark=square*, mark size=2] coordinates {(10,0.077)(25,0.078)(50,0.083)(75,0.081)};
\addlegendentry{$\text{MAE}_g$ ($\downarrow$)}
\end{axis}
\end{tikzpicture}
\caption{Effect of the fixed station-network ratio on NSP performance (fold~3, fixed 50\% context split within each retained station subset). Unlike the earlier context-ratio sweep, this experiment first fixes the available gauge network and then partitions only that network into context and held-out targets. (a)~Radar-evaluated metrics remain stable across 10--75\% retained stations. (b)~Gauge-evaluated metrics are shown for completeness, but their values are not directly comparable to the earlier sweeps because the evaluated station subset changes with the retained network.}
\label{fig:fixed_station_density}
\end{figure}

\paragraph{Fixed station-network ratio.}
A complementary protocol first retains a random fraction of the full station network, then applies the standard 50\% context split within that subset.
Figure~\ref{fig:fixed_station_density} shows the results.
Radar-evaluated metrics remain stable across 10--75\% retained stations, indicating that the degradation in the context-ratio sweep stems from how densely the model conditions on a given network, not from the absolute station count.

Taken together, the three sweeps isolate the source of the accuracy--coherence trade-off: it is the conditioning density relative to the decoder's receptive field, rather than the global gauge count or the training distribution, that governs whether the model preserves spatially coherent fields or collapses toward point-wise gauge interpolation.

\FloatBarrier

\subsection{Gauge context ablation}
\label{app:gauge_ablation}

\begin{figure}[!t]
  \centering
  \includegraphics[width=\linewidth]{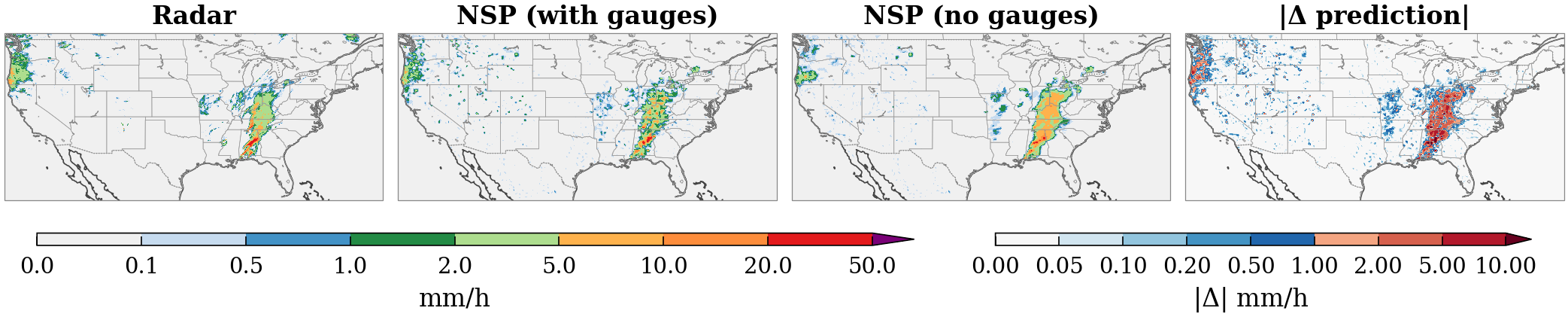}
  \caption{Gauge context ablation for the widespread convective event shown in Figure~\ref{fig:qualitative} (03:00\,UTC, March~16, 2025). Removing gauge context degrades $\text{RMSE}_r$ by 16.7\%; the $|\Delta\,\text{prediction}|$ panel confirms that gauge corrections concentrate in active precipitation regions.}
  \label{fig:gauge_ablation_dep}
  \end{figure}

\begin{figure}[!t]
  \centering
  \includegraphics[width=\linewidth]{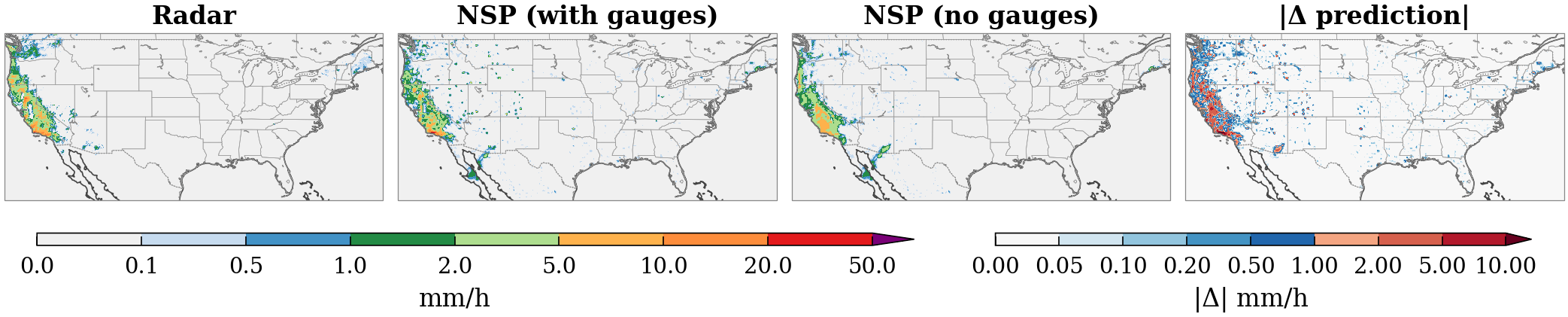}
  \caption{Gauge context ablation for a localized winter precipitation event (13:00\,UTC, December~24, 2025). Removing gauge context changes $\text{RMSE}_r$ by only 1.5\%; the spatial structure is predominantly satellite-driven.}
  \label{fig:gauge_ablation_sat}
  \end{figure}

\begin{table}[!t]
  \centering
  \small
  \caption{Inference latency and model size. Measured on a single NVIDIA H200 GPU with batch size 1, excluding data loading and host--device transfer.}
  \label{tab:compute}
  \setlength{\tabcolsep}{5pt}
  \renewcommand{\arraystretch}{1.1}
  \begin{tabular}{@{} l l r r @{}}
  \toprule
  \textbf{Method} & \textbf{Category} & \textbf{Latency (ms)} & \textbf{Parameters} \\
  \midrule
  GSMaP GC          & Operational product & 0.04 & --- \\
  Linear regression & Statistical         & 1.05 & --- \\
  IDW               & Interpolation       & 1.88 & --- \\
  \textbf{NSP}      & Deep learning       & 2.64 & 4.19M \\
  \bottomrule
  \end{tabular}
  \end{table}

To visualize how gauge observations shape the spatial precipitation field, we compare NSP's output with and without gauge context on two contrasting test cases.
Each panel shows: the radar reference, NSP with gauge context (standard inference), NSP without gauge context (zero gauges), and the absolute difference $|\Delta\,\text{prediction}|$ between the two outputs.

\paragraph{Gauge-dependent case.}
Figure~\ref{fig:gauge_ablation_dep} shows the same widespread convective event as Figure~\ref{fig:qualitative} (03:00\,UTC, March~16, 2025; 9\% rain coverage, peak 50\,mm/h).
Removing all gauge stations degrades $\text{RMSE}_r$ by 16.7\% (0.930 $\to$ 1.085\,mm/h).
The $|\Delta|$ panel reveals that gauge corrections concentrate in active precipitation regions (mean $|\Delta|{=}1.26$\,mm/h in rainy pixels vs.\ 0.04\,mm/h in dry pixels, a 31$\times$ ratio), confirming that the model uses gauge information primarily for intensity refinement rather than spatial structure.

\paragraph{Satellite-driven case.}
Figure~\ref{fig:gauge_ablation_sat} shows a localized winter precipitation event (13:00\,UTC, December~24, 2025; 7\% rain coverage, peak 22\,mm/h).
Despite comparable gauge availability, removing all stations degrades $\text{RMSE}_r$ by only 1.5\% (0.594 $\to$ 0.603\,mm/h).
The $|\Delta|$ panel still shows a 37$\times$ rainy-to-dry correction ratio (0.96 vs.\ 0.03\,mm/h), but the overall impact on accuracy is negligible---the spatial structure is almost entirely satellite-driven.

Together, these two cases illustrate that NSP adaptively balances satellite and gauge information: gauge context provides meaningful intensity corrections for complex events, while the model remains robust when gauges contribute little new information.

\FloatBarrier

\vspace{-2mm}
\subsection{Computational cost}
\label{app:compute}
\vspace{-1mm}

Table~\ref{tab:compute} compares single-sample inference latency and model size for NSP and representative baselines (single NVIDIA H200, batch size 1, fold~3 checkpoint).
NSP requires 2.64\,ms per $260 \times 590$ grid in FP16---a single hourly CONUS field is processed in under 3\,ms, well within real-time operational constraints.
The overhead relative to statistical baselines (IDW, linear regression) is 1--2\,ms, a negligible cost given the substantial accuracy gains reported in Section~\ref{sec:results}.
NSP's 4.19M parameters (3.29M encoder, 825K decoder, 82K SDE) are compact compared to typical U-Net or Vision Transformer architectures used in weather prediction~\citep{pathak2022fourcastnet, bi2023pangu}, where parameter counts routinely exceed 100M.

\FloatBarrier

\vspace{-2mm}
\subsection{Conditional DDPM baseline}
\vspace{-1mm}
\label{app:ddpm}

\begin{figure}[!t]
  \centering
  \textbf{Success case~1} (13:00\,UTC, February~11, 2024)\\[2pt]
  \includegraphics[width=\linewidth]{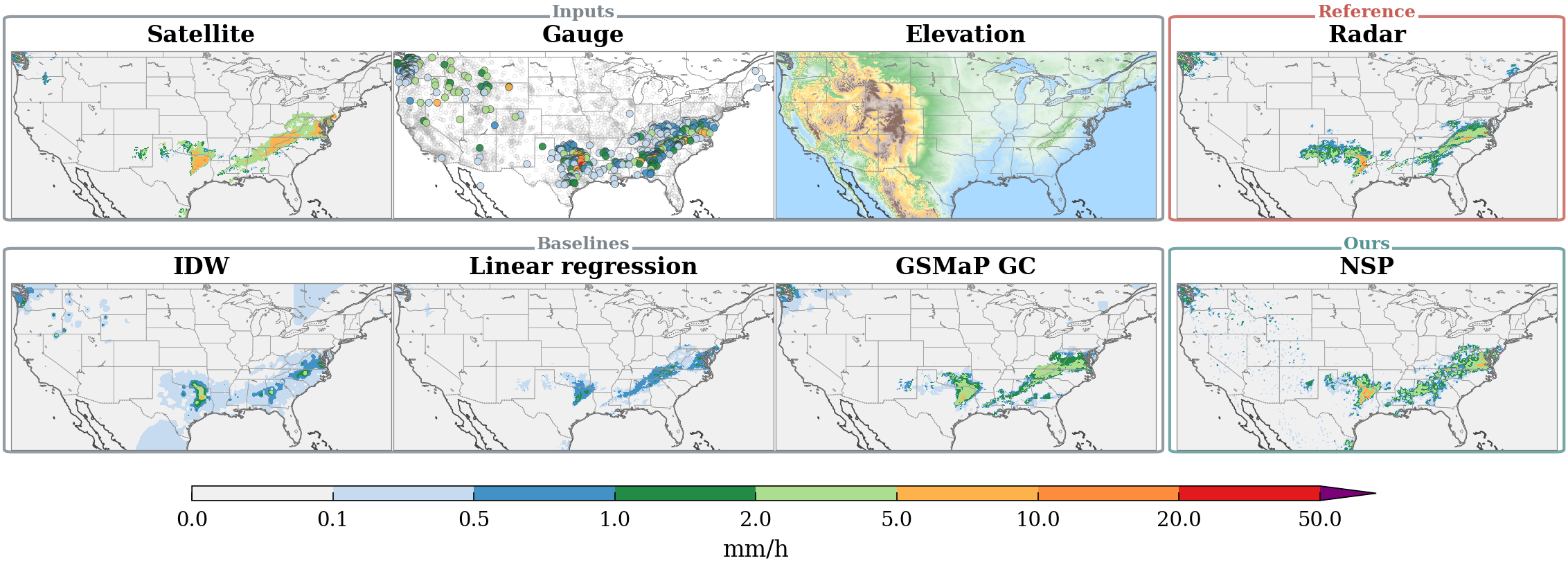}\\[1pt]
  \includegraphics[width=\linewidth]{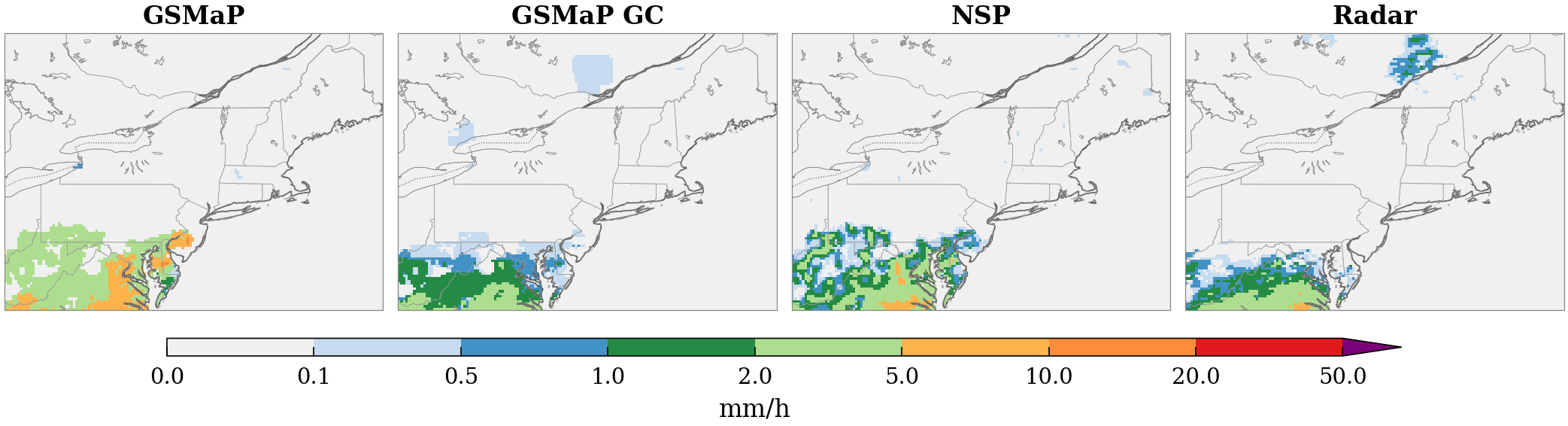}\\[6pt]
  \textbf{Success case~2} (13:00\,UTC, April~5, 2023)\\[2pt]
  \includegraphics[width=\linewidth]{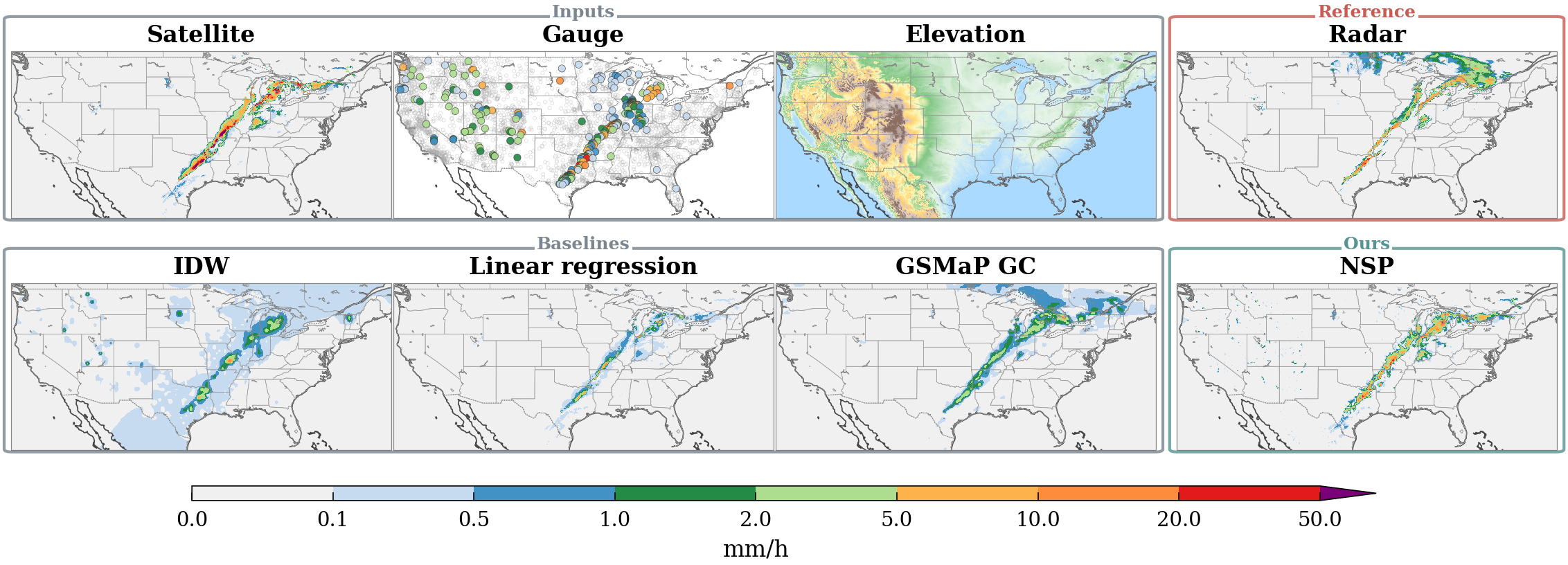}\\[1pt]
  \includegraphics[width=\linewidth]{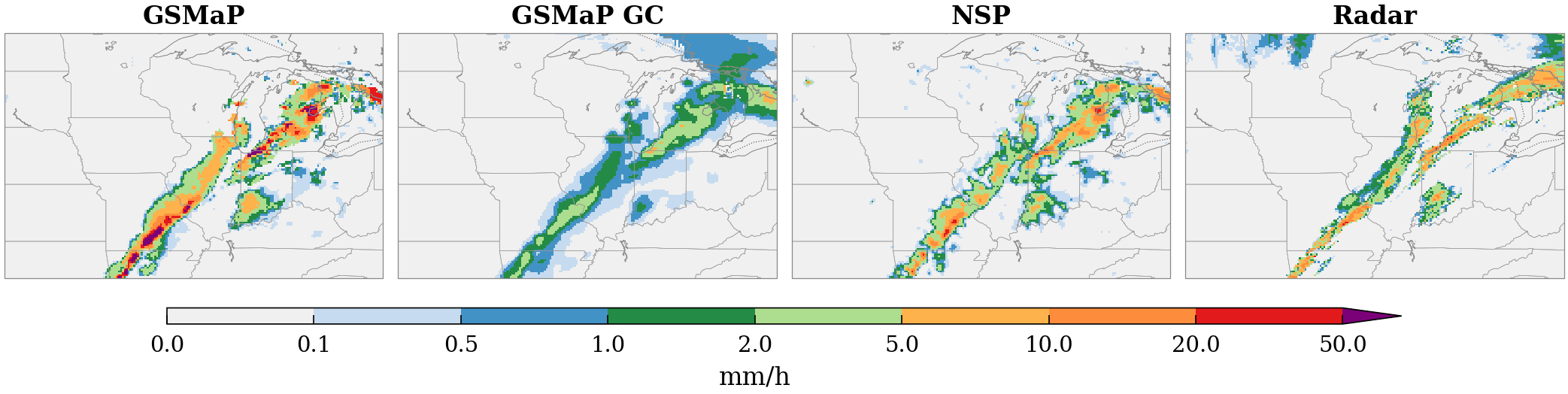}
  \vspace{-2mm}
  \caption{Two success cases, each showing a full CONUS view (top) and regional zoom (bottom).
  \textbf{Case~1:}~A warm-front system over the southeastern US---GSMaP overestimates the spatial extent; GSMaP~GC attenuates but introduces discontinuities; NSP produces a coherent field matching radar and preserves the frontal band orientation.
  \textbf{Case~2:}~An intense squall line from Texas to the Great Lakes---GSMaP inflates the convective core far beyond the radar boundary; NSP recovers the sharp leading edge and trailing stratiform gradient.}
  \label{fig:app_success}
\end{figure}

\begin{figure}[!t]
  \centering
  \includegraphics[width=\linewidth]{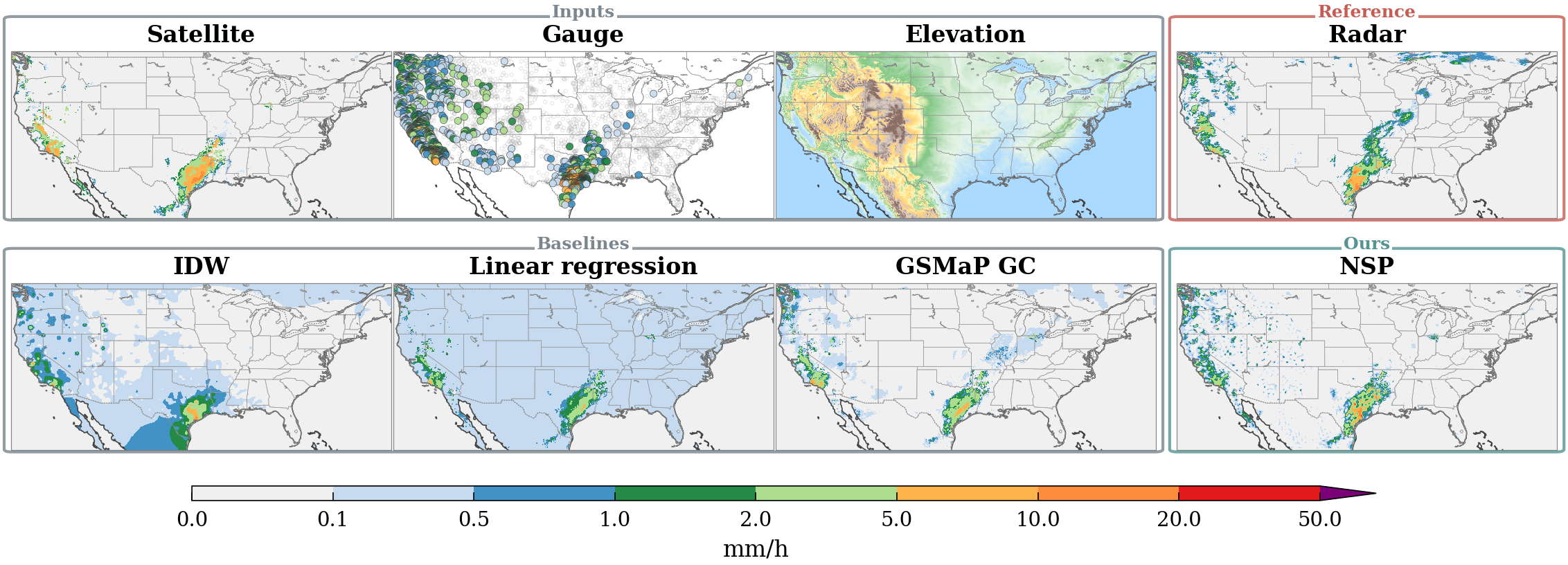}\\[1pt]
  \includegraphics[width=\linewidth]{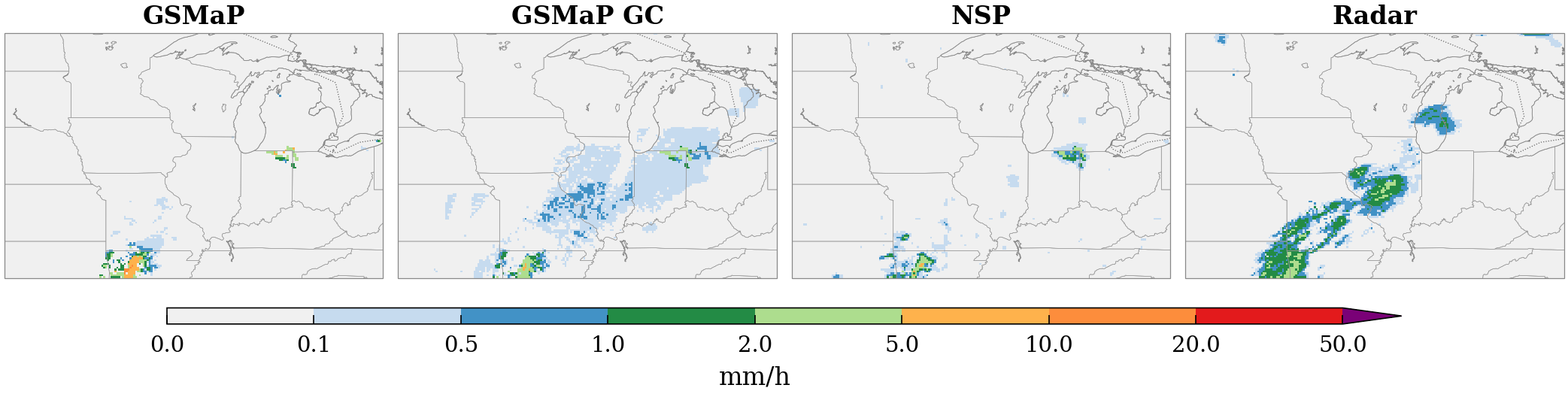}
  \vspace{-2mm}
  \caption{Failure case at 13:00\,UTC on January~22, 2024.
  \textbf{Top:}~Full CONUS---radar detects moderate precipitation over the upper Midwest, but the satellite input is near-zero; without a spatial prior, NSP's residual correction cannot reconstruct the field.
  \textbf{Bottom:}~Regional zoom---GSMaP fails to detect shallow warm-rain precipitation; GSMaP~GC produces diffuse, incoherent patches; NSP, dependent on the satellite residual, cannot recover the missing structure.}
  \label{fig:app_failure}
\end{figure}

\begin{table}[!t]
\centering
\small
\caption{Conditional DDPM baseline vs.\ NSP on CONUS (mean $\pm$ std over three-fold cross-validation). The DDPM performs far worse than NSP on every metric, with large cross-fold variance. All metrics follow the same protocol as Table~\ref{tab:performance_comparison_refined}.}
\label{tab:ddpm}
\setlength{\tabcolsep}{4pt}
\renewcommand{\arraystretch}{1.15}
\begin{tabular}{@{} l cccccc @{}}
\toprule
\textbf{Method} & $\text{RMSE}_r \downarrow$ & $\text{MAE}_r \downarrow$ & $\text{RMSE}_g \downarrow$ & $\text{MAE}_g \downarrow$ & $r_{r,\text{coll}} \uparrow$ & $\text{FSS}_R \uparrow$ \\
\midrule
DDPM & 7.482{\scriptsize $\pm$2.534} & 6.441{\scriptsize $\pm$2.365} & 8.178{\scriptsize $\pm$2.814} & 7.605{\scriptsize $\pm$2.555} & 0.361{\scriptsize $\pm$0.161} & 0.060{\scriptsize $\pm$0.039} \\
\rowcolor{highlightblue}
\textbf{NSP (Ours)} & \textbf{2.818}{\scriptsize $\pm$0.062} & \textbf{1.444}{\scriptsize $\pm$0.026} & \textbf{0.393}{\scriptsize $\pm$0.047} & \textbf{0.076}{\scriptsize $\pm$0.013} & \textbf{0.478}{\scriptsize $\pm$0.021} & \textbf{0.527}{\scriptsize $\pm$0.022} \\
\bottomrule
\end{tabular}
\end{table}

We additionally explored a conditional DDPM as a generative baseline to test whether diffusion-based refinement could outperform deterministic approaches in our sparse-gauge setting.
The DDPM was conditioned on the same four-channel input as all other learned baselines and trained with standard $\epsilon$-prediction on the held-out gauge targets following the shared protocol of Appendix~\ref{app:baseline_loss}.

Table~\ref{tab:ddpm} compares the DDPM against NSP.
The DDPM's mean $\text{RMSE}_r$ of 7.482 is nearly twice that of the weakest main-table baseline (Quantile mapping, 4.07), and its cross-fold standard deviation exceeds 2.5 on the error metrics ($\text{RMSE}_r$, $\text{RMSE}_g$, $\text{MAE}_r$, $\text{MAE}_g$), indicating that training did not converge to a stable solution across data splits.

This instability likely stems from the mismatch between the diffusion training objective and the final evaluation setting: $\epsilon$-prediction optimizes a noise-matching loss in a multi-step denoising process, but evaluation is performed on single-step rain-rate predictions at sparse held-out gauge locations.
We therefore treat this DDPM experiment as a diagnostic negative result rather than a stable baseline, and did not include it in the main comparison.

\section{Additional Qualitative Results}
\label{app:qualitative}

We present two additional success cases and one failure case to complement the qualitative analysis in Section~\ref{sec:results}.

\paragraph{Success case~1: warm-front stratiform precipitation (southeastern US).}
Figure~\ref{fig:app_success} (top) shows a warm-front event on February~11, 2024, producing widespread stratiform precipitation from the Tennessee Valley to the mid-Atlantic coast.
GSMaP overestimates the spatial extent with a broad swath of moderate intensity extending well beyond the radar-observed envelope.
IDW reproduces gauge locations as isolated patches, destroying spatial coherence.
GSMaP~GC suppresses the satellite overestimation but introduces discontinuous patches that do not reflect the underlying frontal structure.
NSP combines the satellite field with gauge corrections to recover a spatially continuous band whose morphology and intensity gradient agree with the radar reference.
The regional zoom reveals the characteristic diffuse overestimation of passive microwave retrieval over stratiform systems, where the ice-scattering signal extends beyond the actual surface precipitation footprint.
GSMaP~GC's daily gauge adjustment overcompensates, attenuating the core of the frontal band, whereas NSP preserves the northeast--southwest orientation and internal intensity gradient.

\paragraph{Success case~2: intense springtime squall line (midwestern US).}
Figure~\ref{fig:app_success} (bottom) captures a vigorous squall line on April~5, 2023, stretching diagonally from Texas to the Great Lakes with peak intensities exceeding 20\,mm/h.
GSMaP dramatically overestimates both the extent and intensity, spreading high values far beyond the actual convective core.
GSMaP~GC narrows the precipitation footprint but still produces a broad, smeared band that lacks the sharp leading edge observed by radar.
NSP recovers a narrow, well-defined squall line morphology with internal intensity gradients consistent with the radar field.
The regional zoom shows that the satellite retrieval inflates the mesoscale convective system several hundred kilometers beyond the radar-observed boundary.
GSMaP~GC corrects the bulk intensity but retains an overly smooth field that obscures the sharp leading-edge structure.
NSP accurately delineates the convective line, reproducing the rapid intensity gradient from the leading edge to the trailing stratiform region---a spatial pattern critical for severe weather nowcasting.

\paragraph{Failure case: satellite detection failure (midwestern US).}
Figure~\ref{fig:app_failure} illustrates a fundamental limitation of residual correction methods.
On January~22, 2024, radar observed moderate precipitation over the upper Midwest (Michigan--Indiana region), yet GSMaP's passive microwave retrieval failed to detect the event, returning near-zero values over the affected area.
Because NSP operates by predicting a residual correction to the satellite estimate (Eq.~\ref{eq:residual}), the absence of a satellite signal leaves no spatial prior to refine: the model cannot reconstruct precipitation structure from sparse gauge observations alone.
The regional zoom confirms that the radar reference shows a coherent precipitation cluster east of Lake Michigan, while the GSMaP panel is nearly blank---a characteristic failure of passive microwave retrieval for shallow, warm-rain systems whose ice-scattering signature is too weak for satellite detection~\citep{kidd2011satellite}.
GSMaP~GC interpolates scattered gauge reports into diffuse low-intensity patches bearing no spatial resemblance to the radar field.
NSP, constrained by the near-zero satellite input, produces minimal output.
This case exposes the inherent ceiling of residual correction frameworks: when the satellite estimate lacks any spatial signal, gauge observations alone---however accurate at point locations---cannot fill the information gap needed to reconstruct a spatially coherent field.
Overcoming this limitation would require an architecture capable of generating precipitation structure de novo from gauge context, a direction for future work.

\end{document}